\documentclass{article}

 \usepackage[preprint]{neurips_2026}

\usepackage[utf8]{inputenc} 
\usepackage[T1]{fontenc}    
\usepackage{hyperref}       
\usepackage{url}            
\usepackage{booktabs}       
\usepackage{amsfonts}       
\usepackage{nicefrac}       
\usepackage{microtype}      
\usepackage{xcolor}         
\usepackage{enumitem}
\usepackage{amsmath}
\usepackage{multirow}
\usepackage{adjustbox}
\usepackage[table]{xcolor}
\usepackage{bm}
\usepackage{makecell}
\usepackage{wrapfig}
\usepackage{subcaption}

\definecolor{base}{HTML}{F7E2A7}
\definecolor{blue}{HTML}{CCE4FF}

\usepackage{algorithm}
\usepackage{algpseudocode}
\usepackage{amssymb}

\newcommand{\bluecomment}[1]{\hfill\textcolor[HTML]{4D81F6}{// #1}}

\definecolor{citecolor}{HTML}{0071bc}
\definecolor{paleplum}{rgb}{0.8, 0.6, 0.8}
\hypersetup{
  colorlinks,
  citecolor=citecolor,
  linkcolor=red
}

\title{Dynamics-Aware Sparse Attention for Efficient Autoregressive Video Diffusion}

\author{%
  \textbf{Yuanyu He$^{1,2*}$ \hspace{0.5em} Zhuokun Chen$^{3*}$ \hspace{0.5em} Yefei He$^{1,\dagger}$ \hspace{0.5em} Zhiwei Tang$^{2}$} \\
  \textbf{Jiasheng Tang$^{2}$ \hspace{0.5em} Yinghao Yu$^{2}$ \hspace{0.5em} Jianfei Cai$^{3}$ \hspace{0.5em} Bohan Zhuang$^{1,\ddagger}$} \\
  \vspace{-0.4em} \\ $^{1}$Zhejiang University \quad $^{2}$Alibaba Group \quad  $^{3}$Monash University
}

\begin{document}

\maketitle

\begingroup
\renewcommand{\thefootnote}{}
\footnotetext[0]{%
  \normalfont
  \ensuremath{*}\,Equal contribution.\quad
  \ensuremath{\dagger}\,Project Lead.\quad
  \ensuremath{\ddagger}\,Corresponding author.%
}
\endgroup

\begin{abstract}

Autoregressive Diffusion Transformers have emerged as a powerful paradigm for long video generation. However, they are often constrained by prohibitive computational overhead and substantial memory footprints. This inefficiency arises from dense computation mechanisms that allocate uniform processing resources across all spatiotemporal regions, failing to leverage the inherent temporal redundancy of video where motion is typically confined to sparse areas. In this paper, we propose \textbf{Dynamics-Aware Sparse Attention (DASA)}, a unified framework that shifts from dense computation to sparse, motion-driven generation to eliminate both storage and computational redundancies. Inspired by video codec principles, our approach explicitly decouples video content into dynamic and static regions. To mitigate memory bottlenecks, we selectively compress the historical KV cache by retaining only dynamic features within the generated chunk. Furthermore, to reduce computational overhead, we employ a region-level computation allocation strategy based on localized dynamics. Additionally, a dynamics-guided distillation method is introduced to enhance performance via a lightweight distillation training process. Experiments on VBench demonstrate that our approach accelerates autoregressive video generation by 2.16$\times$, while preserving high visual fidelity. Code is available on our project page: \url{https://yuanyu0.github.io/DASA/}.

\end{abstract}

\section{Introduction}

Video generation has witnessed unprecedented breakthroughs, propelled by the rapid progress of both proprietary~\cite{brooks2024sora,kling25,seedance2025seedance15,seedance2026seedance2} and open-source~\cite{wan2025wan,kong2024hunyuanvideo,wu2025hunyuanvideo15,lin2024open} foundational models. While these models demonstrate remarkable spatial-temporal fidelity, their outputs remain largely constrained to brief 5- to 20-second clips due to the quadratic computational complexity inherent in full-attention mechanisms. To achieve causally consistent, minute-level synthesis, the field is shifting toward Autoregressive (AR) Diffusion Models~\cite{hyworld2025,lingbot-world,bruce2024genie,wang2026matrix,xie2025progressive}. By factorizing the generation process into sequential chunks, AR diffusion offers a promising pathway toward infinite-length video synthesis.

Despite these compelling advantages, chunk-wise AR diffusion currently faces two severe efficiency bottlenecks in long video generation. First, prohibitive memory expansion. To maintain long-term causal consistency, AR video models typically retain the full historical KV cache across chunks. For example, when generating a 480P 60-second video with LingBot-World~\cite{lingbot-world}, the KV cache can occupy 300GB of memory space, while the model weights only require 56GB. Moreover, the KV cache grows further as video resolution and duration increase. Second, severe latency stems from redundant computation. Current AR architectures blindly allocate identical dense compute resources to every token per chunk, despite most scenes having high temporal redundancy.

\begin{figure}[t]
  \centering
  \includegraphics[width=\textwidth]{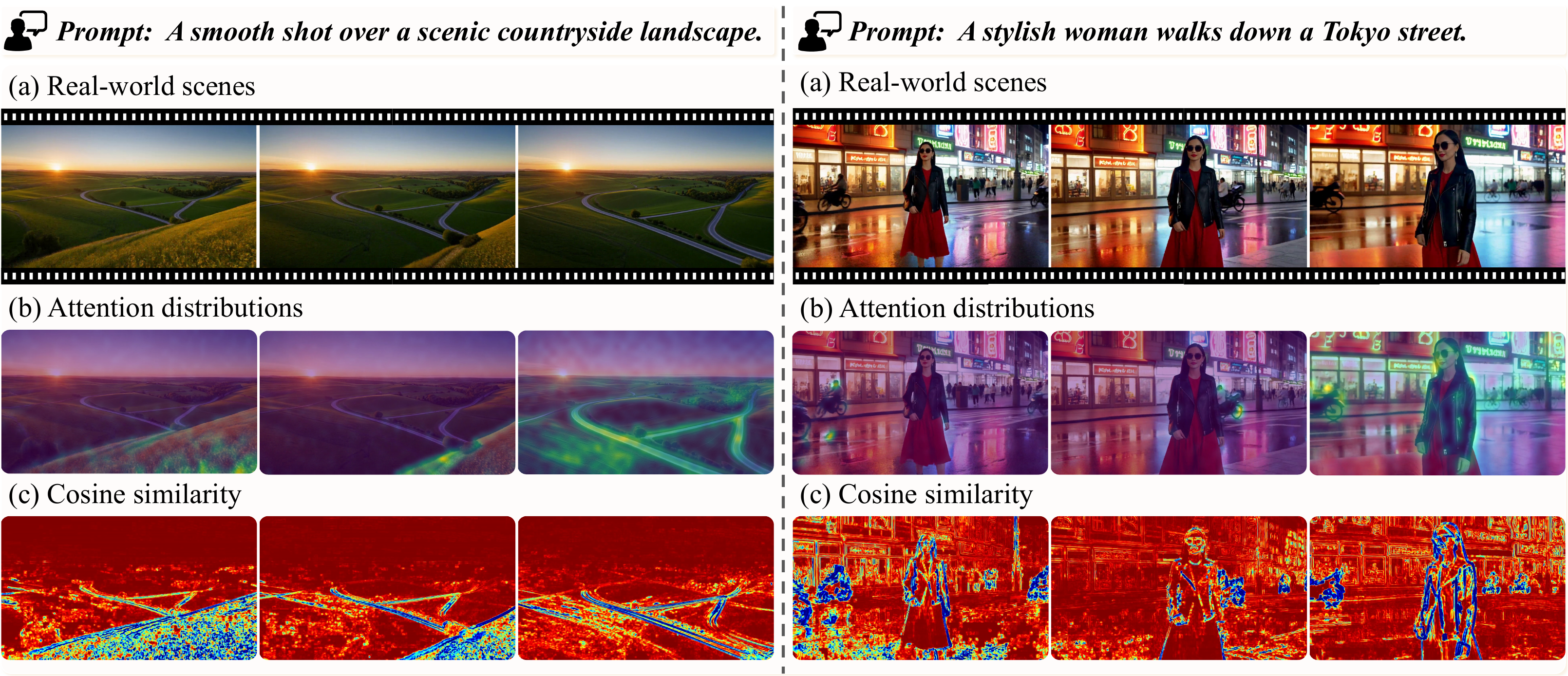}
\vspace{-0.2cm}
\caption{Spatiotemporal dynamic analysis of video generation under different motion priors (left: camera panning; right: subject motion).
\textbf{(a)}~\textbf{Real-world scenes.} The videos are synthesized via Self Forcing, conditioned on a single text prompt provided at the initialization stage.
\textbf{(b)}~\textbf{Attention distributions.} Visualizations reveal the distribution of temporal attention scores, indicating that the generation process predominantly focuses on active dynamic elements alongside adjacent temporal context.
\textbf{(c)}~\textbf{Cosine similarity} between the currently generated frame and preceding frames. The consistently high similarity observed in static background regions demonstrates the potential for extensive feature reuse during computation.
}
\label{fig:motivation}
\end{figure}

%
%
While recent acceleration methods~\cite{guo2026efficient,lv2026light,xu2026sparse} for AR diffusion have analyzed attention patterns to prune inactive tokens or heads, they often overlook the inherent nature of video—where most scenes exhibit high temporal redundancy and motion is confined to sparse regions. 
As shown in Figure~\ref{fig:motivation} (a), real-world videos are dominated by static backgrounds and continuous dynamics, demonstrating massive spatiotemporal redundancy in adjacent frames. 
Addressing these intertwined memory and computational bottlenecks is therefore crucial for efficient AR-diffusion based video generation.

To address these challenges, we propose \textbf{Dynamics-Aware Sparse Attention (DASA)}, an efficient framework for AR video generation. Inspired by modern video coding principles~\cite{zhang2025dense,sarkar2026cope}, we tackle both memory and computational bottlenecks through a unified dynamic and static separation perspective. From a storage standpoint, our analysis reveals that AR models primarily attend to historical dynamic elements and immediate temporal contexts (Figure~\ref{fig:motivation} (b)). Based on this observation, we introduce a Frame-wise Memory Management module that employs layer-wise similarity pruning and a static token refresh strategy. This allows the framework to aggressively compress the KV cache while preserving essential high-frequency details. From a computational perspective, we implement a Region-level Computation Allocation strategy. By identifying static regions that exhibit minimal feature variation across chunks (Figure~\ref{fig:motivation} (c)), we utilize an early-exit mechanism to bypass redundant forward passes, focusing resources exclusively on dynamic updates. By integrating these mechanisms, our framework curtails the KV cache memory footprint while reducing the computational overhead during chunk-wise video generation.

While these mechanisms are inherently training-free, pursuing higher acceleration ratios and framework scalability necessitates their integration into the training phase. However, directly exposing the network to static-feature reuse during training creates an optimization shortcut, resulting in motion sluggishness. To counteract this degradation, we introduce a Dynamics-guided Distillation strategy. By leveraging motion priors, we apply a spatially varying loss reweighting that explicitly amplifies the learning penalty in dynamic regions, successfully preserving the natural dynamics of the generated videos.

In summary, our main contributions are as follows:

    
    


\begin{itemize} [leftmargin=*]
    \item We identify that current autoregressive video diffusion suffers from inefficient resource allocation across highly redundant sequences. We observe that the generation process predominantly focuses on active dynamic elements, while static backgrounds exhibit extensive feature redundancy across adjacent chunks.
    
    \item We propose Dynamics-Aware Sparse Attention (DASA) to eliminate spatiotemporal redundancy from both storage and computation perspectives. By integrating adaptive memory management, motion-aware computation skipping, and dynamics-guided distillation, DASA facilitates scalable video synthesis while avoiding static degradation.
    
    \item Extensive experiments demonstrate DASA's effectiveness in accelerating autoregressive video synthesis while preserving high visual fidelity, achieving a 2.16$\times$ inference speedup at 1080P with a minimal quality drop of less than 0.2\%.
\end{itemize}

\section{Related Work}

\subsection{Video Diffusion Models}

The pursuit of ultra-long and interactive video generation~\cite{ji2026videoar,huang2026live,henschel2025streamingt2v,chen2023seine,cui2026lol,yuan2026helios,chen2026contextforcingconsistentautoregressive,yi2025deep,chen2025sana,hong2025relic,li2025stable,team2025inferix} has been fundamentally shaped by the struggle against computational and memory bottlenecks. Early video diffusion models~\cite{wan2025wan,kong2024hunyuanvideo,zheng2024open,hong2022cogvideo} relied on full bidirectional attention to generate short clips, but the quadratic complexity of attention mechanisms imposed a strict ceiling on video length. To circumvent this, the community shifted toward autoregressive diffusion~\cite{gao2025ca2,gu2025long,zhai2025stargen,sun2025ar}, which decomposed image-to-video generation into either frame-wise~\cite{zhang2025framepack,weng2024art} or chunk-wise~\cite{ai2025magi1autoregressivevideogeneration} steps. Approaches like Causvid~\cite{yin2025causvid} apply DMD~\cite{yin2024improved} to distill 50-step diffusion model into a 4-step generator, enabling faster streaming generation. However, scaling this autoregressive paradigm for long-term consistency requires caching extensive historical tokens, which inevitably triggers an unsustainable memory explosion. Consequently, recent models~\cite{huang2025self,yang2025longlive} often employ a sliding window strategy as a trade-off to maintain a fixed-length memory. Simultaneously, to address the exposure bias in long rollouts, various training-inference alignment mechanisms were proposed, including Diffusion-Forcing~\cite{yin2025causvid}, Self-Forcing~\cite{huang2025self}, Rolling-Forcing~\cite{liu2025rolling}, and Causal-Forcing~\cite{zhu2026causal}. Currently, this trajectory has culminated in interactive world models~\cite{bruce2024genie,wang2026matrix,lingbot-world,hyworld2025,zhang2025epona}. For these world models, efficiency is no longer merely about generating faster, but rather a strict prerequisite to achieving real-time interaction
, necessitating a fundamental rethinking of how foundational models allocate computational resources.

\subsection{Efficient Autoregressive Video Generation}

To meet the stringent efficiency demands of AR video generation, recent efforts have explored various acceleration strategies, which broadly fall into two directions. The first direction focuses on altering the generation paradigm or introducing auxiliary modules to reduce computational latency. For instance, MAGI-1~\cite{ai2025magi1autoregressivevideogeneration} breaks sequential bottlenecks by enabling the parallel generation of multiple video chunks with varying noise levels. Other works focus on context compression. For example, Framepack~\cite{zhang2025framepack} compresses historical frames using fixed patterns based on frame-wise importance, while LoViC~\cite{jiang2025lovic} introduces an autoencoder to compress historical contexts into compact representations.

The second direction accelerates inference by exploiting the model's intrinsic properties, predominantly through attention sparsification. Specifically, Dummy Forcing~\cite{guo2026efficient} prunes inactive attention heads during generation. Light Forcing~\cite{lv2026light} introduces a progressive, chunk-aware sparsity allocation and hierarchical sparse attention to capture information in a coarse-to-fine manner. Furthermore, Sparse Forcing~\cite{xu2026sparse} integrates sparsity into the training phase, motivated by the empirical observation that attention concentrates on salient visual blocks. Yet, it primarily optimizes storage and computation on the KV cache side, lacking optimization for the current query computation.

Fundamentally, while existing approaches primarily derive acceleration strategies from model-centric observations like attention distributions or structural priors~\cite{xi2025sparse,zhang2025fast,zhang2025spargeattention,yang2025sparse,xu2025xattention,he2025neighboring,he2026tc,tang2026kforcingjointnextktokendecoding,xia2025training,zhang2025training}, they often overlook the intrinsic physical nature of the video data itself. We argue that a more natural source of efficiency lies in the data-centric characteristics: video motion is inherently sparse and highly concentrated, whereas the background remains largely static.

\section{Method}

\begin{figure}[t]
  \centering
  \includegraphics[width=\textwidth]{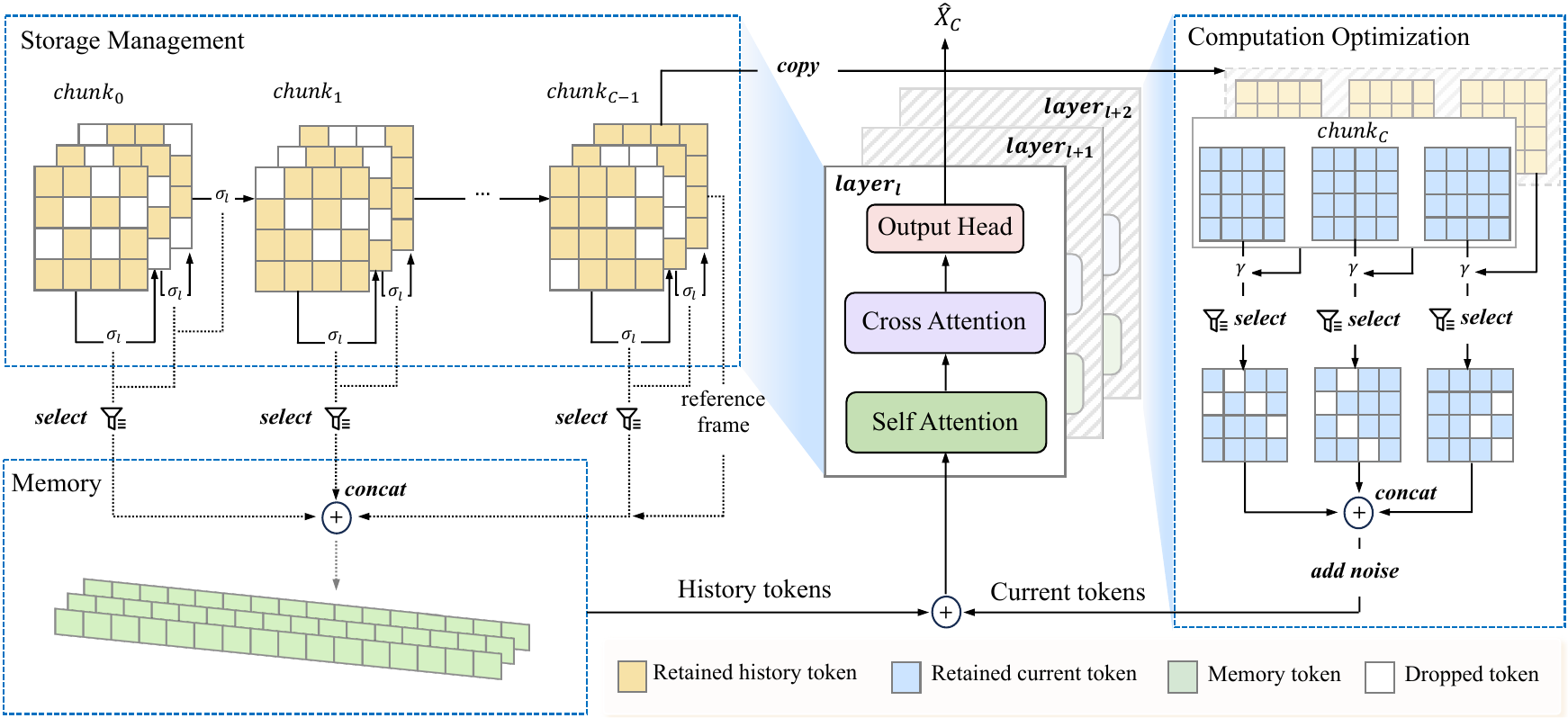}
\caption{\textbf{Overall architecture of the proposed framework.} We introduce a unified approach to optimize AR diffusion generation for both storage and computational efficiency.
\textbf{Storage Management (Left):} Tokens from historical chunks are selectively cached in memory based on a layer-wise threshold $\sigma_l$, discarding redundant information. The final frame of the immediately preceding chunk serves as a \textit{reference frame}, with all its tokens unconditionally preserved. 
\textbf{Computation Optimization (Right):} This reference frame acts as an anchor to guide token selection for the current $\text{chunk}_C$. Redundant tokens are filtered out via a threshold $\gamma$ to reduce computational overhead.
\textbf{AR Diffusion Generation (Middle):} Finally, the retained historical tokens from memory and the active tokens from the current chunk are concatenated and processed by the AR diffusion.
}
\label{fig:framework}
\vspace{-0.2cm}
\end{figure}

To address the redundancy inherent in long video generation, we present Dynamics-Aware Sparse Attention, a unified framework that jointly optimizes AR video generation from the perspectives of memory storage and computational allocation. The key idea is to exploit the temporal redundancy of video by explicitly distinguishing dynamic regions. We first introduce our Frame-wise Memory Management in Sec~\ref{sec:memory}, which alleviates memory bottlenecks by selectively retaining dynamic features in the historical KV cache, together with a static refresh mechanism to maintain long-range consistency. We then describe Region-level Computation Allocation in Sec~\ref{sec:computation}, which reduces redundant computation by assigning different processing budgets to static and dynamic regions within each generated chunk. Finally, in Sec~\ref{sec:training}, we incorporate these sparse generation strategies into training through Dynamics-Guided Distillation Adaptation, a lightweight distillation procedure that mitigates the potential static bias introduced by sparsification and improves robustness across generative settings. Figure~\ref{fig:framework} provides an overview of the proposed framework.

\subsection{Frame-wise Memory Management}
\label{sec:memory}

Due to the high spatial-temporal redundancy in continuous video streams, storing all tokens indiscriminately not only incurs severe memory footprint but is also fundamentally unnecessary. Current AR video models typically employ a sliding window approach~\cite{yin2025causvid, huang2025self}, sometimes paired with sink frames~\cite{yang2025longlive}, to constrain memory by retaining only recent context. However, rigidly discarding older memory creates severe temporal inconsistencies in interactive world generation, especially when navigating back to past scenes. To overcome this limitation, rather than simply dropping old frames, we propose a Frame-wise Memory Management module, which dynamically regulates token caching through similarity-based pruning and a static token refresh strategy. We provide more algorithm details in Appendix ~\ref{app:memory}.

\textbf{Layer-wise retention ratio.} Motivated video coding standards like H.264, which utilize Intra-coded and Predictive-coded frames, we determine whether a token should be retained in memory based on its feature similarity with its adjacent frame. Specifically, let $x^k_{t}$ denote the latent representation of the $k$-th spatial token at frame $t$. We compute the cosine similarity between temporally adjacent tokens:
\begin{equation}
    \mathrm{sim}_t^k = \cos\left(\bm{x}_t^k, \bm{x}_{t+1}^k \right).
\end{equation}
Unlike H.264, which uses the previous frame as a reference, our approach retains the last generated frame as the reference frame and save all its tokens. This strategy results in significantly smoother generated videos. We hypothesize that this entire frame serves as an anchor for the newly generated chunk, which aligns with our observations in Figure~\ref{fig:motivation} (b). If a token’s representation exhibits minimal changes compared to its latter frame, it is deemed redundant and can be safely discarded from the memory buffer. Thus, a preliminary retention mask is generated by thresholding the similarity: $M_{\text{sim}}^{k} = \mathbb{I} \left( \text{sim}_t^{k} < \sigma \right)$, where $\mathbb{I}(\cdot)$ is the indicator function and $\sigma$ is the threshold.

As visualized in the Appendix ~\ref{app:attn_dist}, we observe that the models attend to more dynamic tokens as the layer goes deeper. We assume that shallow layers capture high-frequency details that change rapidly, while deeper layers extract abstract, low-frequency semantic information that remains highly stable across frames. Leveraging this property, we apply a layer-wise retention rate by linearly decaying the threshold $\sigma_l$ for layer $l$ as the layer depth increases:
\begin{equation}
    \sigma_l = \sigma_{max} - \frac{\sigma_{max} - \sigma_{min}}{L} \cdot l
\end{equation}
where $\sigma_{max}$ and $\sigma_{min}$ are the predefined upper and lower bounds of the threshold, and $L$ is the total number of layers. A lower threshold in deeper layers enforces a stricter retention condition, further compressing the memory footprint.

\textbf{Static token refresh strategy.} While similarity-based pruning significantly reduces redundancy, we observe temporal degradation in static regions as they are continuously pruned. If a background token remains similar across multiple frames and is repeatedly dropped, the model is forced to rely on a heavily outdated token. Consequently, the model fails to capture subtle temporal variations—such as global illumination shifts—leading to error accumulation that severely degrades overall visual fidelity.

To mitigate this, we introduce a Static Token Refresh strategy. We maintain an aging counter $\Delta t^k$ for each spatial position $k$, defined as the number of frames elapsed since a token was last written to the memory at its location: $\Delta t^k = t - t_{\text{write}}^k$. If a token at a specific location has been continuously dropped for a predefined maximum age threshold $\tau$, we force the model to retain this token. The final boolean mask $M_{final}^k$, which determines whether the $k$-th token is saved to the memory bank, is defined as the union of the similarity and aging conditions:
\begin{equation}
    M_{\text{final}}^k = M_{\text{sim}}^k \lor \mathbb{I} \left(\Delta t^k \ge \tau \right).
\end{equation}
When $M_{\text{final}}^k = 1$, the token is stored, and its timestamp $t_{\text{write}}^k$ is reset to the current frame $t$. This mandatory refresh acts as a periodic calibration, effectively preventing the fading of high-frequency background details while keeping the overall memory growth well-controlled.

\subsection{Region-level Computation Allocation}
\label{sec:computation}

Unlike traditional video diffusion models~\cite{wan2025wan,kong2024hunyuanvideo} that compute full spatial-temporal attention across the entire video sequence simultaneously, AR diffusion models adopt a chunk-wise generation paradigm. While this can efficiently scale to long video generation, we observe a significant computational bottleneck: the standard diffusion process iteratively denoises all tokens within a chunk, ignoring the strong inter-chunk temporal coherence. In reality, the current generating chunk shares substantial similarities with the preceding synthesized chunk, particularly in backgrounds and smoothly moving subjects. To effectively eliminate this redundant computation, we propose a \textit{Region-level Computation Allocation} strategy that dynamically reuses previously computed features from the preceding chunk, thereby skipping unnecessary computations for highly redundant regions. More algorithm details are provided in Appendix ~\ref{app:allocation}.

\textbf{Motion-aware early exit.} In the first denoising step of the current chunk $C$, the model predicts an estimated clean state $\hat{X}_C$. We leverage this intermediate $\hat{X}_C$ to evaluate temporal redundancy against the previously finalized chunk $X_{C-1}$. Since physical motion in videos is typically local and continuous, we perform a local alignment search rather than comparing strictly identical spatial coordinates. 

Specifically, for a predicted token $\hat{\bm{x}}_C^{(i,j)}$ at spatial coordinate $(i, j)$ in chunk $C$, we search for its best matching token within a local $w \times w$ spatial window $\mathcal{N}_w(i,j)$ in the previous chunk $C-1$. The aligned token $\tilde{\bm{x}}_{C-1}^{(i,j)}$ and its corresponding maximum similarity score $\mathrm{sim}_C^{(i, j)}$ are defined as:
\begin{equation}
    \tilde{\bm{x}}_{C-1}^{(i,j)} = \bm{x}_{C-1}^{(u^*,v^*)}, \quad \text{where} \quad (u^*,v^*) = \mathop{\arg\max}_{(u,v) \in \mathcal{N}_w(i,j)} \mathrm{cos} \left( \hat{\bm{x}}_C^{(i,j)}, \bm{x}_{C-1}^{(u,v)} \right),
\end{equation}
and $\mathrm{sim}_C^{(i, j)} = \mathrm{cos} \big( \hat{\bm{x}}_C^{(i,j)}, \tilde{\bm{x}}_{C-1}^{(i,j)} \big)$. This local search efficiently tracks moving objects, while static backgrounds naturally align at the exact same coordinate $(i, j)$. Furthermore, by keeping the window size $w$ as a small fixed constant, this local search avoids the quadratic $\mathcal{O}(L^2)$ complexity of a global matching process (where $L$ is the spatial sequence length). Instead, it scales linearly with $L$, ensuring the additional alignment overhead remains minimal.

Based on the alignment results, a token is classified as redundant if its maximum similarity exceeds a predefined threshold $\gamma$. This yields a binary mask $M$:
\begin{equation}
    M^{(i,j)} = \mathbb{I} \left( \mathrm{sim}_C^{(i, j)} \ge \gamma \right).
\end{equation}

For these redundant tokens, we halt their subsequent denoising iterations—triggering an early exit—and directly copy the corresponding aligned tokens from the previous chunk. Let $\tilde{X}_{C-1}$ denote the feature map composed of all aligned tokens. The output latent $\hat{X}_{C}^\text{out}$ is updated via spatial blending:
\begin{equation}
    \hat{X}_{C}^{\text{out}} = M \odot \tilde{X}_{C-1} + (1 - M) \odot \mathcal{F}_\theta(\hat{X}_C)
\end{equation}
where $\mathcal{F}_\theta(\cdot)$ denotes the rest of the denoising forward passes applied exclusively to the remaining unaligned (dynamic) tokens. This streamlined process securely bypasses heavy, unnecessary forward passes for visually redundant areas without compromising generation quality.

\begin{wrapfigure}{r}{0.55\textwidth}
  \centering
  \includegraphics[width=0.53\textwidth]{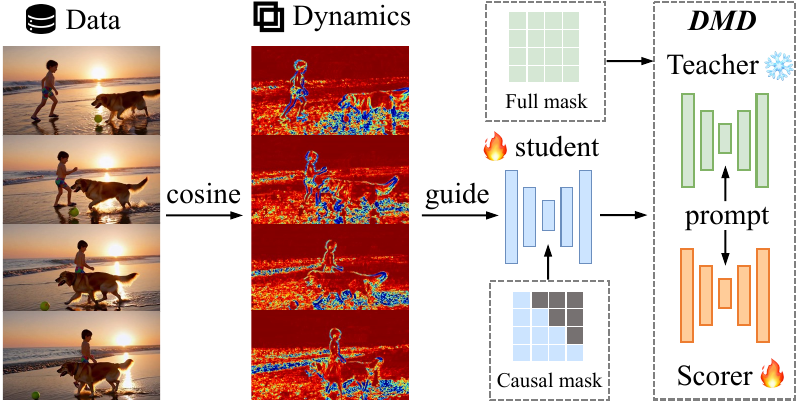}
  \caption{\textbf{Dynamics-guided Distribution Matching Distillation.} To mitigate static bias, we introduce a dynamic weight mask based on temporal cosine similarity. By amplifying loss in high-motion regions, it penalizes static predictions and forces the model to learn complex dynamics.}
  \label{fig:architecture}
\end{wrapfigure}

\subsection{Dynamics-Guided Distillation Adaptation}
\label{sec:training}

While our inference technique efficiently accelerates generation and maintains overall quality, minor domain shifts or artifacts can still occur. To further elevate visual quality, we integrate the aforementioned mechanisms into the training phase. However, during this native adaptation, we find that while the spatial quality of the generated videos improves, they suffer from motion lag, causing actions and object movements to become noticeably restricted.

We attribute this degradation to an optimization shortcut. When the region-level static scene reuse mechanism is exposed during training, the network discovers a trivial solution to minimize the reconstruction error. Since predicting static or minimal changes yields a safely lower loss than taking the risk to generate complex motions, the model develops a ``static bias.'' It over-relies on copying historical tokens and inadvertently suppresses its internal capability to generate dynamic motions.

To counteract this static bias without discarding the efficiency gains, we introduce a \textit{Dynamics-Guided Distillation Adaptation} built upon the standard Distribution Matching Distillation~\cite{yin2024one} framework. We leverage the robust motion priors from the data sample to explicitly penalize static predictions. Specifically, let $\bm{z}_t^{(i,j)}$ denote the token representation of the sample video at frame $t$ and spatial coordinate $(i,j)$. We compute the temporal similarity $\text{sim}_t^{(i,j)}$ between adjacent frames using the previously defined cosine similarity. To identify areas of significant change and penalize ``lazy'' predictions, we map this similarity into a spatially-varying dynamic weight mask $W_t^{(i,j)}$:
\begin{equation}
    W_t^{(i,j)} = 1 + \alpha \cdot \max \left(0, 1 - \text{sim}_t^{(i,j)} \right),
\end{equation}
where $\alpha$ is a dynamic scaling factor. 

Finally, this weight mask is applied token-wise to the standard DMD loss $\mathcal{L}_{\text{base}}$. The proposed dynamics-guided distillation loss $\mathcal{L}_{\text{dynamic}}$ is formulated as:
\begin{equation}
    \mathcal{L}_{\text{dynamic}} = \frac{1}{N} \sum_{t,i,j \in \Omega} W_t^{(i,j)} \cdot \mathcal{L}_{\text{base}}^{(t,i,j)},
\end{equation}
where $N$ denotes the total number of tokens. Regions exhibiting larger temporal variations in the sampled data naturally yield a lower similarity $\text{sim}_t^{(i,j)}$, thereby receiving a significantly amplified loss weight. This dynamic guidance acts as a strong penalty against the Student model's optimization shortcut, forcing it to faithfully reconstruct the real data's motion dynamics in the generated videos.

\section{Experiments}

\subsection{Experimental Setup}

\textbf{Implementation details.} We integrate our approach into representative baselines, including Self-Forcing \cite{huang2025self} and LongLive \cite{yang2025longlive}, and further finetune them to enhance sample quality and generation speed. To comprehensively validate our method, our experimental scope spans multiple settings: standard 5s video generation, 30s long video generation, and high-resolution (780P and 1080P) video generation. All efficiency experiments are conducted on NVIDIA H100 GPUs. Due to page limit, we provide more implementation
details in the Appendix~\ref{app:imp_detail}.

\textbf{Evaluation.} We evaluate our method on VBench \cite{huang2024vbench}, a comprehensive video generation benchmark. Specifically, we adopt the extended version of its 946 evaluation prompts introduced by~\cite{huang2025self}. Following standard practice, we sample 5 videos per prompt and assess the quality score and semantic score across all 16 quantitative dimensions defined in VBench. All evaluation experiments are conducted on NVIDIA H20 GPUs.

\begin{table*}[ht]
\centering
\caption{
Quantitative results for \textbf{5-second short video generation} on VBench~\cite{huang2024vbench}. The best results are in \textbf{bold}, and the second best are \underline{underlined}. The color box \colorbox{base!40}{\phantom{d}} denotes the baseline model.
}
\label{short-video}
\begin{adjustbox}{max width=\textwidth}
\begin{tabular}{c l c c c c c c}
\toprule
\multirow{2}{*}{\begin{tabular}[c]{@{}c@{}}Type\end{tabular}} & \multirow{2}{*}{Model} & \multirow{2}{*}{Params} & \multicolumn{2}{c}{Efficiency} & \multicolumn{3}{c}{Visual Quality} \\
\cmidrule(lr){4-5} \cmidrule(lr){6-8}
& & & Throughput (FPS)$\uparrow$ & Speedup$\uparrow$ & Quality Score$\uparrow$ & Semantic Score$\uparrow$ & Total Score$\uparrow$ \\
\midrule
\multirow{5}{*}{\begin{tabular}[c]{@{}c@{}} Diffusion \end{tabular}} & LTX-Video~\cite{HaCohen2024LTXVideo} & 1.9B & 8.98  & - & 82.30 & 70.79 & 80.00 \\
& CogVideoX~\cite{yang2024cogvideox} & 5B & 0.85 & - & 82.68 & 79.48 & 82.04 \\
& Wan2.1~\cite{wan2025} & 1.3B & 0.78 & - &85.30 & 80.09 & 84.26 \\
& Open-Sora~\cite{zheng2024open} & 11B & 0.08 & - & 84.28 & 79.89 & 83.40 \\
& Hunyuan Video~\cite{kong2024hunyuanvideo} & 13B & 0.22 &- &85.07 & 76.88 & 83.43 \\
\midrule
\multirow{15}{*}{\begin{tabular}[c]{@{}c@{}} Autoregressive \end{tabular}} & SkyReels-V2~\cite{chen2025skyreelsv2infinitelengthfilmgenerative} & 1.3B & 0.49 & - & 84.70 & 74.53 & 82.67 \\
& MAGI-1~\cite{ai2025magi1autoregressivevideogeneration} & 4.5B & 0.19 & - & 82.04 & 67.74 & 79.18 \\
& NOVA~\cite{deng2024nova} & 0.6B & 0.88 & - & 80.39 & 79.05 & 80.12 \\
& Pyramid Flow~\cite{jin2024pyramidal} & 2B & 6.77 & - & 84.74 & 69.62 & 81.72 \\
& CausVid~\cite{yin2025causvid} & 1.3B & 0.69 & - & 83.73 & 78.49 &82.69 \\
\cmidrule(lr){2-8}
& \cellcolor{base!40}Self Forcing~\cite{huang2025self} & \cellcolor{base!40}1.3B & \cellcolor{base!40}18.42 & \cellcolor{base!40}1.00$\times$ & \cellcolor{base!40}84.97 & \cellcolor{base!40}81.06 & \cellcolor{base!50}84.19 \\
& TeaCache~\cite{liu2025timestep} & 1.3B & 21.69 & 1.18$\times$ & 84.22 & 80.47 & 83.47 \\
& Dummy Forcing~\cite{guo2026efficient} & 1.3B & 25.19 & 1.37$\times$ & 84.47 & \underline{81.08} & 83.79 \\
& DASA & 1.3B & \underline{26.21} & \underline{1.42$\times$} & \underline{84.57} & 80.79 & \underline{83.81} \\
& DASA w/ adaptation & 1.3B & \textbf{27.58} & \textbf{1.50$\times$} & \textbf{84.64} & \textbf{81.32} & \textbf{83.98} \\
\cmidrule(lr){2-8}
& \cellcolor{base!40}LongLive~\cite{yang2025longlive} & \cellcolor{base!40}1.3B & \cellcolor{base!40}18.30 & \cellcolor{base!40}1.00$\times$ & \cellcolor{base!40}83.77 & \cellcolor{base!40}80.73 & \cellcolor{base!40}83.16 \\
& TeaCache~\cite{liu2025timestep} & 1.3B & 21.82 & 1.19$\times$ & 83.10 & 80.73 & 82.63 \\
& Dummy Forcing~\cite{guo2026efficient} & 1.3B & 25.10 & 1.37$\times$ & 83.16 & \textbf{80.99} & \underline{82.73} \\
& DASA & 1.3B & \underline{26.50} & \underline{1.45$\times$} & \underline{83.18} & \underline{80.87} & 82.72 \\
& DASA w/ adaptation & 1.3B & \textbf{27.91} & \textbf{1.53$\times$} & \textbf{83.78} & 80.35 & \textbf{83.09} \\
\bottomrule
\end{tabular}
\end{adjustbox}
\end{table*}

\subsection{Main Results}

\subsubsection{Short Video Generation}

This subsection evaluates our method on standard 5-second video generation using the VBench dataset, as presented in Table~\ref{short-video}. We integrate DASA into state-of-the-art AR diffusion models and compare it with TeaCache~\cite{liu2025timestep} and Dummy Forcing~\cite{guo2026efficient}. Since modern AR diffusion models typically use a low-step generation paradigm of 4 denoising steps, cross-step caching methods like TeaCache face inherent acceleration limits and yield a marginal speedup of 1.18$\times$. While Dummy Forcing improves this by dropping attention heads based on the first generated chunk, its static pruning pattern may lead to information loss and degraded visual quality.

Unlike these baselines that rely on structural or step-wise approximations, our method exploits the inherent spatial-temporal redundancy of videos to dynamically allocate computational resources based on content complexity. By adaptively assigning more resources to highly dynamic regions while bypassing static areas, our adaptive pipeline achieves a speedup of 1.53$\times$, consistently preserving the high visual and semantic quality. Furthermore, it is worth noting that our method optimizes AR diffusion models at a different level, making it orthogonal and compatible with TeaCache and Dummy Forcing.

\begin{table*}[ht]
\centering
\caption{
Quantitative results for \textbf{30-second long video} and \textbf{high-resolution video} generation on VBench~\cite{huang2024vbench}.
}
\label{long-high-video}
\begin{adjustbox}{max width=0.9\textwidth}
\begin{tabular}{l c c c c c}
\toprule
\multirow{2}{*}{Model} & \multicolumn{2}{c}{Efficiency} & \multicolumn{3}{c}{Visual Quality} \\
\cmidrule(lr){2-3} \cmidrule(lr){4-6}
& Throughput (FPS)$\uparrow$ & Speedup$\uparrow$ & Quality Score$\uparrow$ & Semantic Score$\uparrow$ & Total Score$\uparrow$ \\
\midrule
SkyReels-V2~\cite{chen2025skyreelsv2infinitelengthfilmgenerative} &0.49 &- & 80.77 & 53.37 & 75.29 \\
FramePack~\cite{zhang2025framepack} & 0.92 & - & 83.61 & 75.32 & 81.95 \\
\midrule
\rowcolor{gray!20}
\multicolumn{6}{c}{\textit{30s Video Generation}} \\
\midrule
\rowcolor{base!40}
Self Forcing & 18.42 & 1.00$\times$ & 84.15 & 80.24 & 83.37 \\
Dummy Forcing  & 25.19 & 1.37$\times$ & \textbf{84.11} &79.18 & 83.12 \\
DASA  & \textbf{26.21} & \textbf{1.42$\times$} & 84.05 & \textbf{79.78} & \textbf{83.20} \\
\cmidrule(lr){1-6}
\rowcolor{base!40}
LongLive & 18.30 & 1.00$\times$ & 83.57 & 80.96 & 83.05 \\
Dummy Forcing  & 25.10 & 1.37$\times$ & 83.23 &79.59 &82.50 \\
DASA  &\textbf{26.50} &\textbf{1.45$\times$} &\textbf{83.39} &\textbf{80.59} &\textbf{82.83} \\
\midrule
\rowcolor{gray!20}
\multicolumn{6}{c}{\textit{720P Video Generation}} \\
\midrule
\rowcolor{base!40}
Self Forcing & 5.92 & 1.00$\times$ & 84.30 & 81.21 & 83.68 \\
Dummy Forcing  & 9.52 & 1.61$\times$ & 84.16 & 80.86 & 83.50 \\
DASA  & \textbf{10.10} & \textbf{1.71$\times$} & \textbf{84.40} & \textbf{80.92} & \textbf{83.71} \\
\midrule
\rowcolor{base!40}
LongLive & 5.98 & 1.00$\times$ & 83.76 & 81.50 & 83.31 \\
Dummy Forcing  & 9.74 & 1.63$\times$ & \textbf{83.49} & \textbf{81.63} & \textbf{83.12} \\
DASA  &\textbf{10.37} &\textbf{1.73$\times$} &83.38 &81.54 & 83.01 \\
\midrule
\rowcolor{gray!20}
\multicolumn{6}{c}{\textit{1080P Video Generation}} \\
\midrule
\rowcolor{base!40}
Self Forcing & 1.35 & 1.00$\times$ & 84.12 & 79.42 & 83.18 \\
Dummy Forcing & 2.64 & 1.96$\times$ & 83.10 & \textbf{79.89} & 82.46 \\
DASA & \textbf{2.81} & \textbf{2.08$\times$} & \textbf{83.55} & 79.73 & \textbf{82.79} \\
\midrule
\rowcolor{base!40}
LongLive & 1.36 & 1.00$\times$ & 82.13 & 80.40 & 81.78 \\
Dummy Forcing  & 2.74 & 2.02$\times$ & \textbf{81.92} &80.24 &81.58 \\
DASA &\textbf{2.94} &\textbf{2.16$\times$} &81.90 & \textbf{80.55} & \textbf{81.63} \\
\bottomrule
\end{tabular}
\end{adjustbox}
\end{table*}

\subsubsection{Long Video Generation}

Crucially, we highlight the scalability of our approach in 30-second long video generation, with quantitative results presented in Table~\ref{long-high-video}. By maintaining contextual coherence over extended frames, DASA seamlessly scales to longer sequences. When integrated with baselines like Self Forcing and LongLive, our method achieves speedups of up to 1.45$\times$, demonstrating its capability to accelerate extended generation without compromising overall semantic integrity.

\begin{figure}[t]
  \centering
  \includegraphics[width=\textwidth]{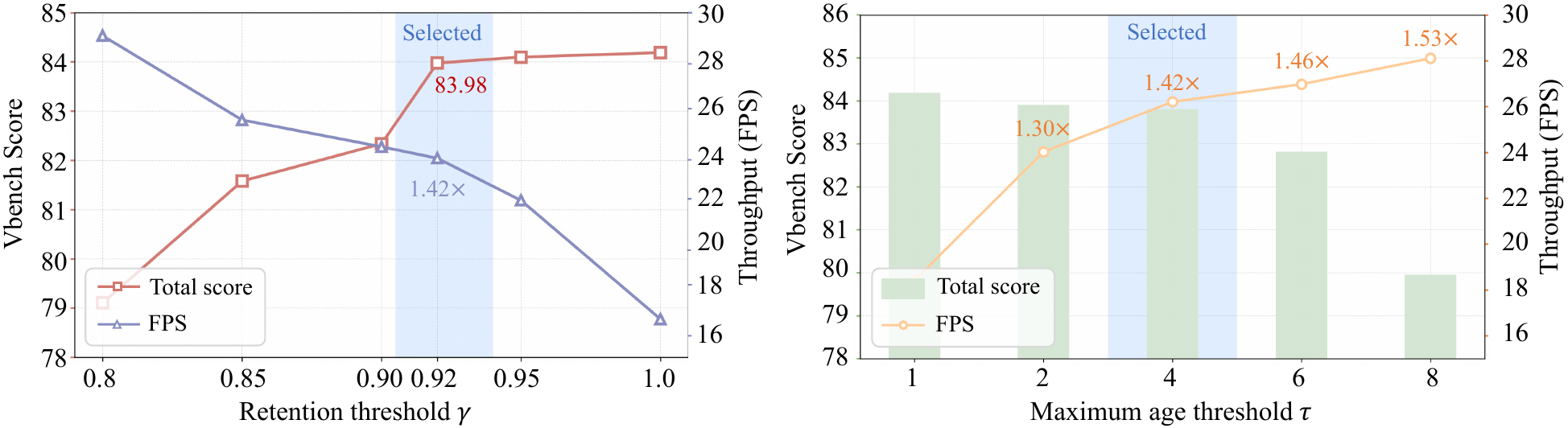}
\caption{Ablation study on the impact of early-exit threshold $\gamma$ and maximum age threshold $\tau$.}
\label{fig:retention_age}
\vspace{-0.2cm}
\end{figure}

\subsubsection{High-resolution Video Generation}

To further validate the robustness of our method under high computational demands, we extend our evaluation to 720P and 1080P high-resolution generation, as detailed in Table~\ref{long-high-video}. Notably, the acceleration benefits of DASA scale with resolution. Our approach delivers speedups of up to 2.16$\times$ at 1080P. Despite massive reductions in memory and computational overhead, DASA successfully preserves superior visual fidelity and fine spatial details, yielding good trade-off between throughput and total video quality.

\subsection{Ablation Study}

\textbf{Effect of memory and computation retention.} We evaluate two thresholds controlling spatial-temporal redundancy: memory retention $\sigma$ and early-exit $\gamma$. To avoid a rigid speed-quality trade-off, we adopt a dynamic layer-wise $\sigma$ (denser in shallow layers, sparser in deep ones), which successfully reduces memory footprint while preserving fidelity, as detailed in Table \ref{retention}. For the Motion-Aware Early Exit, Figure~\ref{fig:retention_age} (a) demonstrates that $\gamma$ dictates the speed-quality balance; an optimal $\gamma=0.92$ achieves a 1.42$\times$ speedup with a competitive Vbench total score of 83.98. In combination, these approaches bypass unnecessary memory operations and computational overhead.

\textbf{Effect of static token refresh.} As shown in the Figure~\ref{fig:retention_age} (b), increasing $\tau$ boosts throughput, but excessively large values ($\tau \ge 6$) severely degrade visual quality. This confirms our premise that unconstrained token dropping causes severe error accumulation. Conversely, the baseline ($\tau=1$) preserves details but limits acceleration. We select $\tau = 4$ as the optimal balance, achieving a $1.42\times$ speedup with highly competitive visual fidelity.

\begin{table}[htbp]
\centering
\begin{minipage}{0.49\textwidth}
\centering
\caption{
Ablation study on the impact of memory retention ratio $\sigma$.
}
\resizebox{\linewidth}{!}{
\label{retention}
\begin{tabular}{c c | c c c c}
\toprule
\multicolumn{2}{c|}{$\sigma$} & \multirow{2}{*}{\makecell{Throughput\\(FPS)$\uparrow$}} & \multirow{2}{*}{\makecell{Quality\\Score$\uparrow$}} & \multirow{2}{*}{\makecell{Semantic\\Score$\uparrow$}} & \multirow{2}{*}{\makecell{Total\\Score$\uparrow$}} \\
\cmidrule(lr){1-2}
$\sigma_\text{max}$ & $\sigma_\text{min}$ & & & & \\
\midrule
\multicolumn{2}{c|}{\cellcolor{base!40}Self Forcing} &\cellcolor{base!40}18.42 &\cellcolor{base!40}84.97 &\cellcolor{base!40}81.06 &\cellcolor{base!40}84.19 \\
0.8 & 0.8 & 22.34 (1.21$\times$) & 84.89 & 81.00 & 84.11 \\
0.6 & 0.6 & 24.11 (1.31$\times$) & 84.12 & 80.04 & 83.30 \\
0.8 & 0.2 & 26.21 (1.42$\times$) & 84.57 & 80.79 & 83.81 \\
\bottomrule
\end{tabular}
}
\end{minipage}
\hfill
\begin{minipage}{0.49\textwidth}
\centering
\caption{
Ablation study on the impact of local alignment and adaptation.
}
\resizebox{\linewidth}{!}{
\label{tab:align_adapt}
\begin{tabular}{c | c c c c}
\toprule
 & \makecell{Throughput\\(FPS)$\uparrow$} & \makecell{Quality\\Score$\uparrow$} & \makecell{Semantic\\Score$\uparrow$} & \makecell{Total\\Score$\uparrow$} \\
\midrule
\cellcolor{base!40}Self Forcing &\cellcolor{base!40}18.42 &\cellcolor{base!40}84.97 &\cellcolor{base!40}81.06 &\cellcolor{base!40}84.19 \\
w/ align. & 26.21 (1.42$\times$) & 84.57 & 80.77 & 83.81 \\
w/ adapt. & 26.32 (1.43$\times$) & 84.70 & 81.24 & 84.01 \\
ours & 27.58 (1.50$\times$) & 84.64 & 81.32 & 83.98 \\
\bottomrule
\end{tabular}
}
\end{minipage}
\end{table}

\textbf{Effect of local alignment.} As shown in Table \ref{tab:align_adapt}, by dynamically identifying and reusing redundant tokens via local window search, our motion-aware early exit bypasses unnecessary denoising tokens, yielding a significant $1.42\times$ throughput increase.

\textbf{Effect of dynamics-guided adaptation.} Introducing the region-reuse mechanism during training can induce a "static bias," restricting motion generation. Our dynamics-guided adaptation explicitly penalizes these static predictions. As shown in Table \ref{tab:align_adapt}, the dynamics-guided adaptation effectively mitigates the motion degradation, achieving the highest Semantic Score by preserving authentic dynamics. Furthermore, this harmonized training boosts the inference throughput to 27.58 FPS ($1.50\times$), validating that our design achieves a desirable balance between efficiency and visual fidelity.

\section{Conclusion}
In this work, we explore the intrinsic dynamics and spatial-temporal redundancies in autoregressive long video generation. To exploit this nature for faster generation, we decompose this redundancy from two orthogonal perspectives: memory footprint and computational overhead. Specifically, we introduce a Frame-wise Memory Management module to dynamically prune temporally redundant tokens and strictly bound KV cache growth. Concurrently, we propose a Region-level Computation Allocation scheme that utilizes a motion-aware early exit mechanism to bypass unnecessary denoising computations for static backgrounds. Extensive experiments demonstrate that our token-level optimizations significantly accelerate video synthesis while maintaining high visual fidelity.

\textbf{Limitations and future work.} While our approach effectively reduces spatial-temporal redundancy at the \textit{token} level, it currently does not explore redundancies across other architectural dimensions. Notably, our method is fundamentally orthogonal and highly compatible with existing state-of-the-art acceleration techniques. For instance, TeaCache exploits redundancies across the \textit{denoising step} dimension, whereas Dummy Forcing accelerates generation from the \textit{attention head} dimension. In future work, we plan to systematically integrate our token-level strategy with step-wise TeaCache and head-wise Dummy Forcing. We anticipate that unifying these three orthogonal dimensions will unlock unprecedented synergistic acceleration ratios for autoregressive video generation.

\newpage
\appendix


\section{Algorithm Details}

This part provides the precise algorithmic formulations of our proposed optimization framework, complementing the conceptual methodology described in the main text. To explicitly demonstrate how we exploit spatial-temporal redundancy for efficient long video generation, we detail the execution pipelines of our two core modules. First, Algorithm \ref{alg:memory_management} formalizes the Frame-wise Memory Management strategy, illustrating the dynamic KV cache pruning and static refresh mechanisms. Second, Algorithm \ref{alg:computation_allocation} outlines the Region-level Computation Allocation scheme, showing how the motion-aware early exit mechanism is practically implemented during the chunk-wise denoising loop.

\subsection{Memory Management}
\label{app:memory}

Algorithm \ref{alg:memory_management} presents the core computational workflow of our Frame-wise Memory Management module, which dynamically optimizes the KV cache for efficient long video generation. The algorithm executes the methodology described in the main text, primarily integrating layer-wise retention threshold decay, similarity-based temporal redundancy pruning, and a mandatory static token refresh strategy. By synergistically applying these mechanisms, our approach reduces memory footprint while effectively preventing the temporal degradation of background details, ensuring high overall visual fidelity.

\begin{algorithm}[H]
\caption{Frame-wise Memory Management for Video Generation}
\label{alg:memory_management}
\begin{algorithmic}[1]
\Require Latent tensor $X \in \mathbb{R}^{B \times N_f \times N_s \times D}$, Keys $K$, Values $V$
\Require Layer index $l$, Total layers $L$, Sim thresholds $\sigma_{\text{max}}, \sigma_{\text{min}}$, Max token age $\tau$
\Require Reference frame $X_{\text{ref}}$, Token age tracker $T_{\text{update}}$, Current frame num $t_{\text{curr}}$
\Ensure Output tensor $O$, Updated KV Cache state
\State \bluecomment{1. Layer-wise Retention Threshold Decay}
\State $\sigma_l \leftarrow \sigma_{\text{max}} - \frac{\sigma_{\text{max}} - \sigma_{\text{min}}}{L} \cdot (l + 1)$
\State \bluecomment{2. Temporal Context Concatenation}
\If{$X_{\text{ref}}$ is not \textbf{null}}
    \State $X_{\text{comb}} \leftarrow \text{Concat}(X_{\text{ref}}, X, \text{dim}=1)$
    \State $K_{\text{comb}} \leftarrow \text{Concat}(K_{\text{ref}}, K, \text{dim}=1)$
    \State $V_{\text{comb}} \leftarrow \text{Concat}(V_{\text{ref}}, V, \text{dim}=1)$
\Else
    \State $X_{\text{comb}} \leftarrow X, \quad K_{\text{comb}} \leftarrow K, \quad V_{\text{comb}} \leftarrow V$
\EndIf
\State \bluecomment{3. Similarity-based Token Pruning}
\State $X_{\text{prev}} \leftarrow X_{\text{comb}}[:, :-N_{\text{ref}}, \dots]$
\State $X_{\text{next}} \leftarrow X_{\text{comb}}[:, 1:(-N_{\text{ref}}+1), \dots]$
\State $\text{sim} \leftarrow \text{CosineSimilarity}(X_{\text{prev}}, X_{\text{next}}, \text{dim}=-1)$
\State $M_{\text{sim}} \leftarrow \text{sim} < \sigma_l$ \bluecomment{Preliminary boolean mask based on similarity}
\State \bluecomment{4. Static Token Refresh Strategy}
\For{each frame index $t \in \{1, \dots, N_f\}$}
    \State $M_{\text{curr}} \leftarrow M_{\text{sim}}[:, t]$
    \State $\Delta t \leftarrow t_{\text{curr}} - T_{\text{update}}$ \bluecomment{Calculate token age}
    \State $M_{\text{age}} \leftarrow \Delta t \ge \tau$
    \State $M_{\text{final}}[:, t] \leftarrow M_{\text{curr}} \lor M_{\text{age}}$ \bluecomment{Force refresh if token is too old}
    \State $T_{\text{update}}[M_{\text{final}}[:, t]] \leftarrow t_{\text{curr}}$ \bluecomment{Reset aging counter for retained tokens}
    \State $t_{\text{curr}} \leftarrow t_{\text{curr}} + 1$
\EndFor
\State \bluecomment{5. Reference Frame Preservation \& Cache Update}
\State $M_{\text{tail}} \leftarrow \mathbf{1} \in \{0, 1\}^{B \times N_{\text{ref}} \times N_s}$ \bluecomment{Unconditionally retain the last frame(s)}
\State $M_{\text{all}} \leftarrow \text{Concat}(M_{\text{final}}, M_{\text{tail}}, \text{dim}=1)$
\State $K_{\text{save}} \leftarrow K_{\text{comb}}[M_{\text{all}}], \quad V_{\text{save}} \leftarrow V_{\text{comb}}[M_{\text{all}}]$ \bluecomment{Extract retained tokens via masking}
\State $\text{KVCache}.\text{Append}(K_{\text{save}}, V_{\text{save}})$
\State $X_{\text{ref}} \leftarrow X[:, -N_{\text{ref}}:, \dots]$ \bluecomment{Update reference frame for the next chunk}
\State \bluecomment{6. Standard Attention Computation with Compact Cache}
\State $O \leftarrow \text{Attention}(Q, \text{KVCache}_K, \text{KVCache}_V)$
\State \Return $O$
\end{algorithmic}
\end{algorithm}

\subsection{Computation Allocation}

\label{app:allocation}

Algorithm \ref{alg:computation_allocation} details the execution pipeline of our Region-level Computation Allocation strategy, specifically highlighting the Motion-Aware Early Exit mechanism. Operating within a chunk-wise generation framework, the algorithm leverages the first denoising step to perform a local spatial-temporal alignment between the current chunk and the preceding finalized chunk. By extracting a binary mask via thresholding and spatial dilation, it accurately segregates dynamic tokens from static or redundant backgrounds. In all subsequent denoising iterations, redundant tokens securely bypass the heavy forward passes (early exit) and are directly blended using cached historical features, thereby significantly eliminating redundant computations while preserving temporal coherence.

\begin{algorithm}[H]
\caption{Region-level Computation Allocation (Motion-Aware Early Exit)}
\label{alg:computation_allocation}
\begin{algorithmic}[1]
\Require Noisy latent chunk $Z_C$, Previous clean chunk $X_{C-1}$
\Require Denoising steps $\{t_0, t_1, \dots, t_{S-1}\}$, Search window size $w$
\Require Similarity threshold $\gamma$, Morphological dilation kernel size $k$
\Ensure Denoised output chunk $X_C$
\State $M_{\text{dyn}} \leftarrow \mathbf{1}$ \bluecomment{Initialize mask to compute all tokens at the first step}
\State $\tilde{X}_{C-1} \leftarrow \text{null}$
\State \bluecomment{1. Spatial Denoising Iterations for Current Chunk $C$}
\For{each step index $s \in \{0, 1, \dots, S-1\}$}
    \State \bluecomment{Execute sparse Generator forward pass for dynamic tokens}
    \State $\hat{X}_C \leftarrow \text{Generator}(Z_C, \text{timestep}=t_s, \text{active\_mask}=M_{\text{dyn}})$
    \State \bluecomment{2. Local Alignment \& Mask Generation (Executed only at step 0)}
    \If{$s = 0$ \textbf{and} $X_{C-1}$ is not \textbf{null}}
        \State \bluecomment{Local motion search within $w \times w$ window to track objects}
        \State $\tilde{X}_{C-1}, S_{\text{sim}} \leftarrow \text{LocalNxNMatch}(X_{C-1}, \hat{X}_C, \text{range}=w)$
        \State \bluecomment{Determine dynamic tokens via max-pooling \& thresholding}
        \State $M_{\text{dyn}} \leftarrow \text{MinPool2D}(S_{\text{sim}}) < \gamma$
        \State $M_{\text{dyn}} \leftarrow \text{Dilate}(M_{\text{dyn}}, \text{kernel}=k)$ \bluecomment{Spatial dilation for dynamic region}
        \State $M_{\text{dyn}} \leftarrow \text{Upsample}(M_{\text{dyn}})$ \bluecomment{Expand to match original sequence length}
        \State $M_{\text{static}} \leftarrow \neg M_{\text{dyn}}$ \bluecomment{Mask for visually redundant regions}
    \EndIf
    \State \bluecomment{3. Spatial Blending \& Computation Early Exit}
    \If{$s > 0$ \textbf{and} $M_{\text{static}}$ is available}
        \State \bluecomment{Bypass computation for static areas, copy from aligned history}
        \State $\hat{X}_C \leftarrow M_{\text{static}} \odot \tilde{X}_{C-1} + M_{\text{dyn}} \odot \hat{X}_C$
    \EndIf
    \State \bluecomment{4. Prepare Latent for Next Denoising Step}
    \If{$s < S-1$}
        \State $Z_C \leftarrow \text{Scheduler.AddNoise}(\hat{X}_C, t_{s+1})$
    \EndIf
\EndFor
\State \bluecomment{Return the fully denoised chunk}
\State $X_C \leftarrow \hat{X}_C$
\State \Return $X_C$
\end{algorithmic}
\end{algorithm}

\section{More Implementation Details}
\label{app:imp_detail}

\textbf{Setup for frame-wise memory management implementation.} Here, we provide the specific implementation details of the proposed Frame-wise Memory Management module. For the layer-wise retention ratio, the similarity threshold decays linearly from an upper bound $\sigma_{max}$ to a lower bound $\sigma_{min}$ across the transformer layers. During the training phase, to enhance the model's robustness and prevent overfitting to strictly static backgrounds, we introduce a stochastic retention mechanism where naturally dropped tokens have a $5\%$ probability of being forcefully retained. For the static token refresh strategy, a token is unconditionally revived if its continuous dropping age exceeds a predefined maximum token age threshold. Furthermore, to ensure seamless temporal transition across autoregressive chunks, we unconditionally preserve the exact last frame of the previously generated chunk to serve as the structural anchor for the current chunk.

\textbf{Setup for computation allocation implementation.} In this section, we detail the engineering setup for the Region-level Computation Allocation and Motion-Aware Early Exit mechanism. During the local alignment search between adjacent chunks, we constrain the spatial search window to a $3 \times 3$ grid to balance matching accuracy and computational overhead. To distinguish redundant tokens, we apply a strict cosine similarity threshold of $\gamma = 0.92$. To conservatively encompass moving subjects and prevent visual artifacts at motion boundaries, the resulting dynamic binary mask undergoes morphological spatial dilation using a $3 \times 3$ kernel (with stride 1 and padding 1). Crucially, to maximize inference efficiency, this motion search and mask generation process is strictly executed \textit{only once} during the first denoising timestep. The acquired mask is then kept fixed for all subsequent denoising steps within the current chunk. Finally, after the spatial denoising loop concludes, an additional forward pass infused with clean context noise is executed to reliably update the KV cache for the upcoming autoregressive chunk.

\section{More Qualitative Results}

\subsection{Attention Distributions}
\label{app:attn_dist}

To validate the motivation behind our layer-wise token retention strategy, we investigate how the model allocates its spatial attention at different stages of generation. To illustrate this progression, we extract and normalize the attention weights from layers 5, 15, and 25, utilizing them as representative examples of shallow, intermediate, and deep network layers, respectively. This visualization helps to confirm our hypothesis regarding the network's transition from capturing fine-grained details to abstract semantic concepts.

\newpage
\begin{figure}[H]
  \centering
  \includegraphics[width=\textwidth]{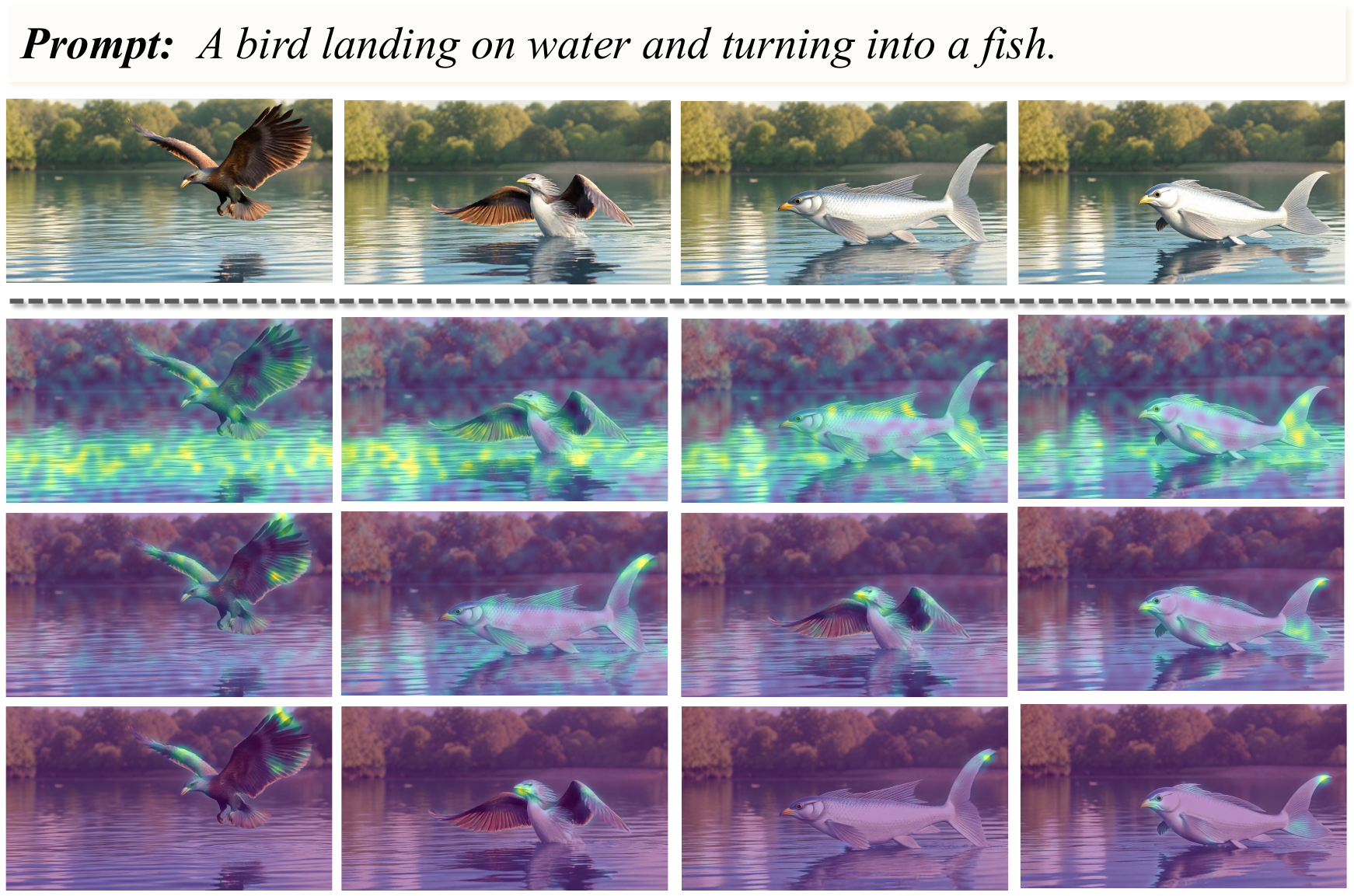}
  \caption{Attention distribution in layers 5, 15, and 25 of LongLive.}
  \label{fig:bird}
\end{figure}

\begin{figure}[H]
  \centering
  \includegraphics[width=\textwidth]{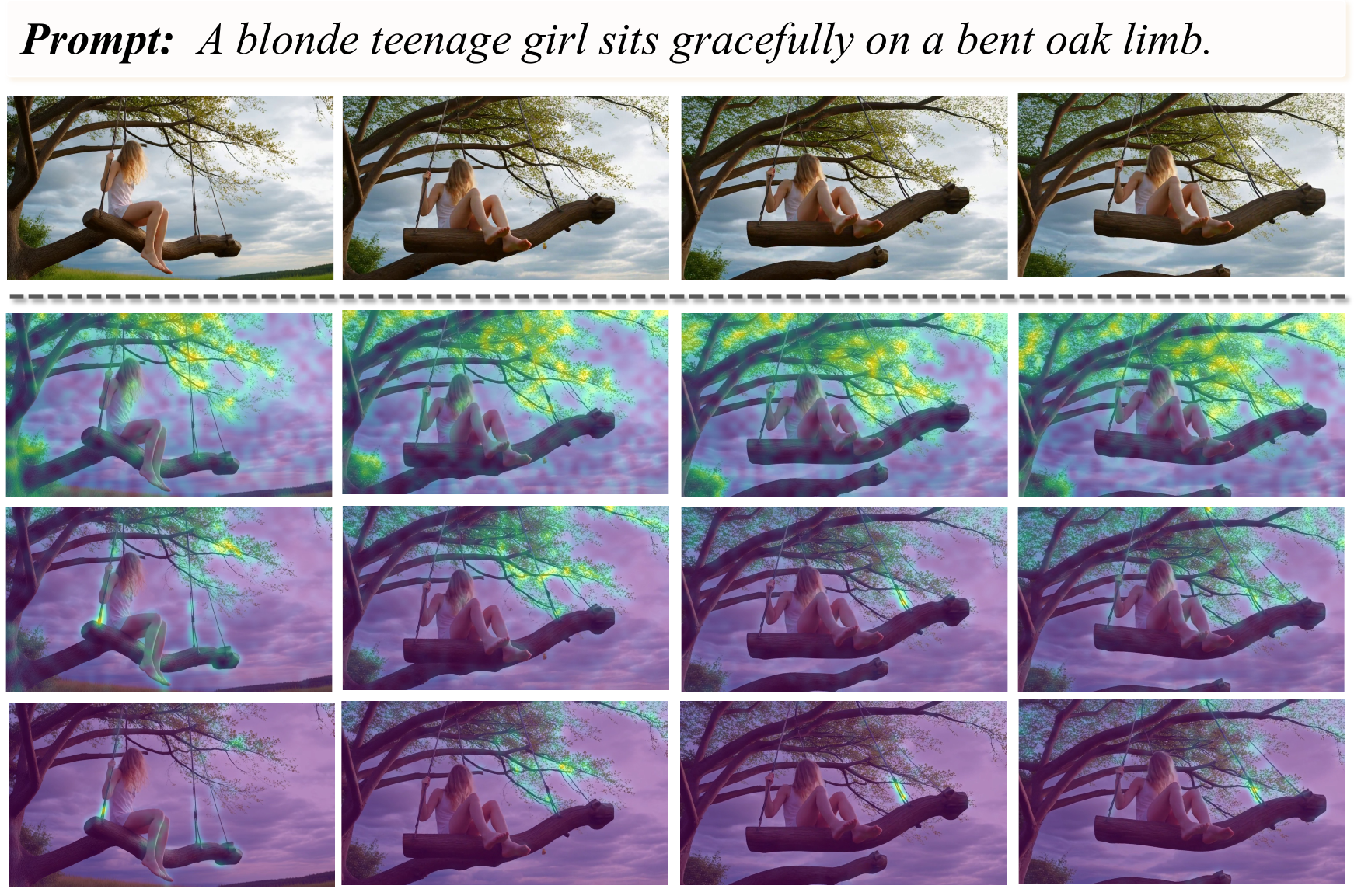}
  \caption{Attention distribution in layers 5, 15, and 25 of LongLive.}
  \label{fig:blond}
\end{figure}

\begin{figure}[H]
  \centering
  \includegraphics[width=\textwidth]{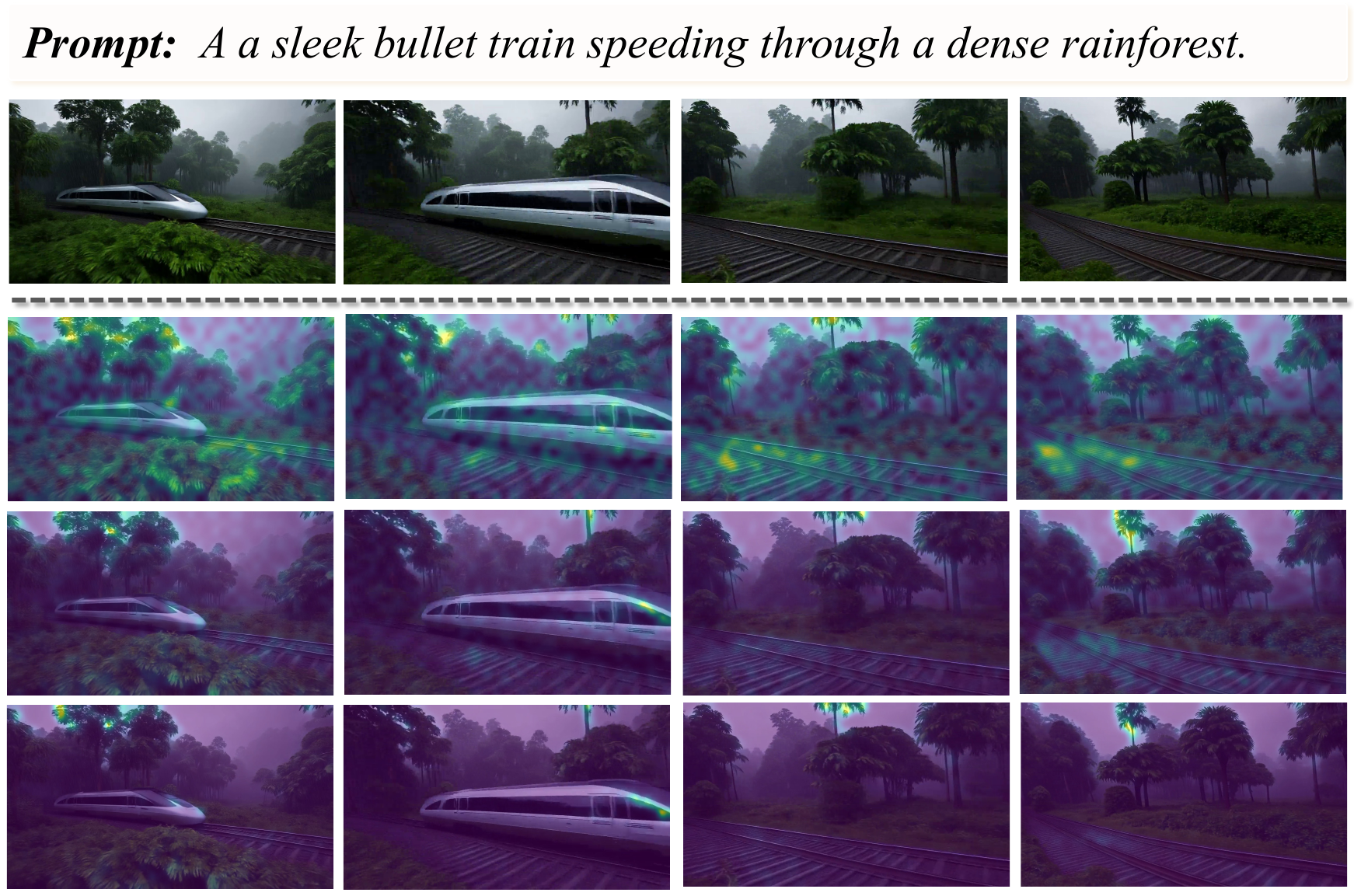}
  \caption{Attention distribution in layers 5, 15, and 25 of LongLive.}
  \label{fig:train}
\end{figure}

\begin{figure}[H]
  \centering
  \includegraphics[width=\textwidth]{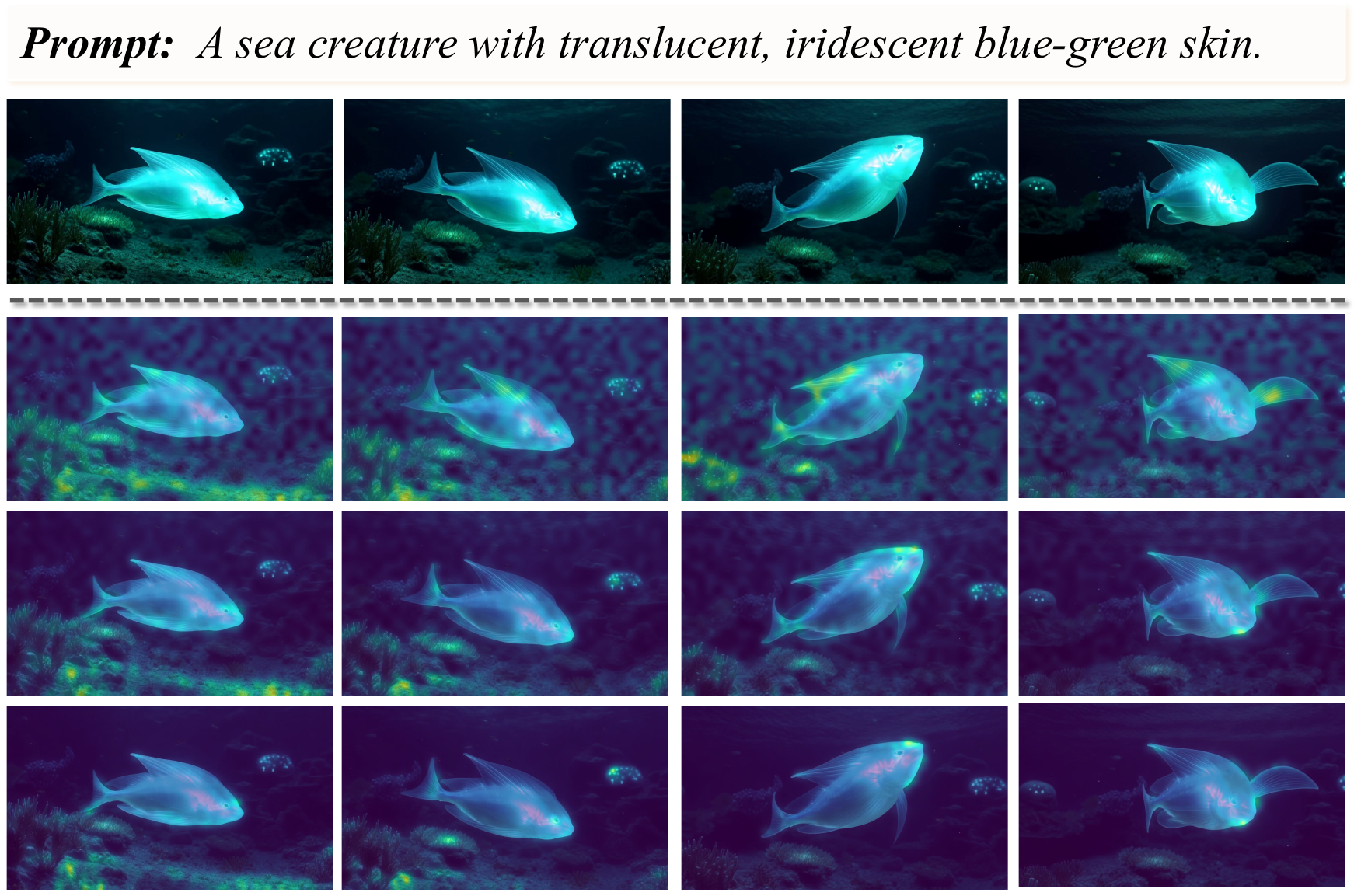}
  \caption{Attention distribution in layers 5, 15, and 25 of LongLive.}
  \label{fig:fish}
\end{figure}

\newpage
\subsection{Visualization}

\vspace{-2cm}

The following qualitative comparison demonstrates that our DASA framework generates video sequences that are visually nearly identical to the Self Forcing. This visual consistency across diverse dynamic scenarios—including fluid human actions, complex camera panning, detailed static backgrounds, and fast-moving vehicles—validates that our spatiotemporal redundancy reduction strategy achieves high generation quality while providing substantial computational acceleration.

\begin{figure}[H]
  \centering
  \includegraphics[width=\textwidth]{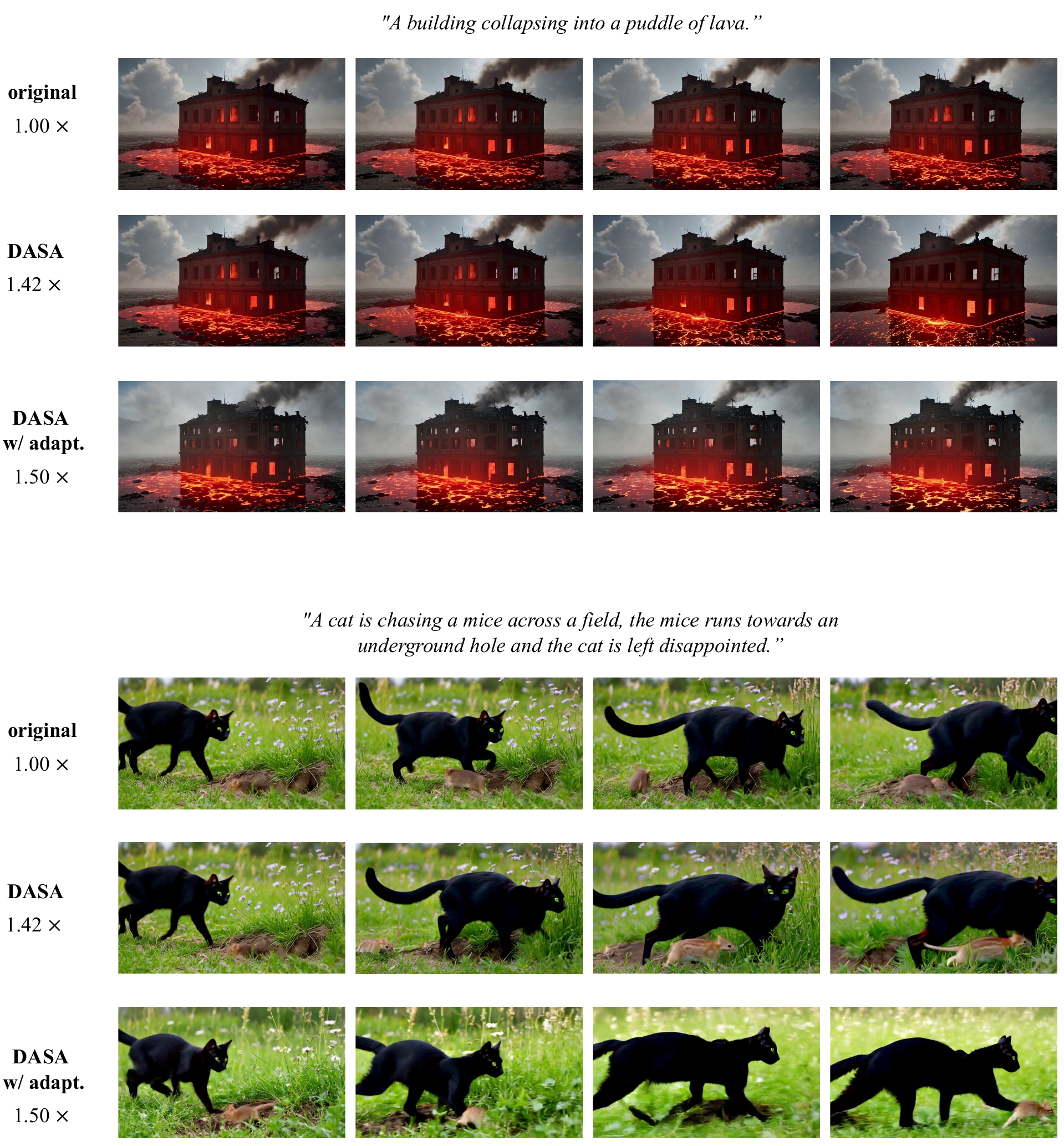}
  \caption{Visualization of 480P videos generated by Self Forcing.}
  \label{fig:fish}
\end{figure}

\begin{figure}[H]
  \centering
  \includegraphics[width=\textwidth]{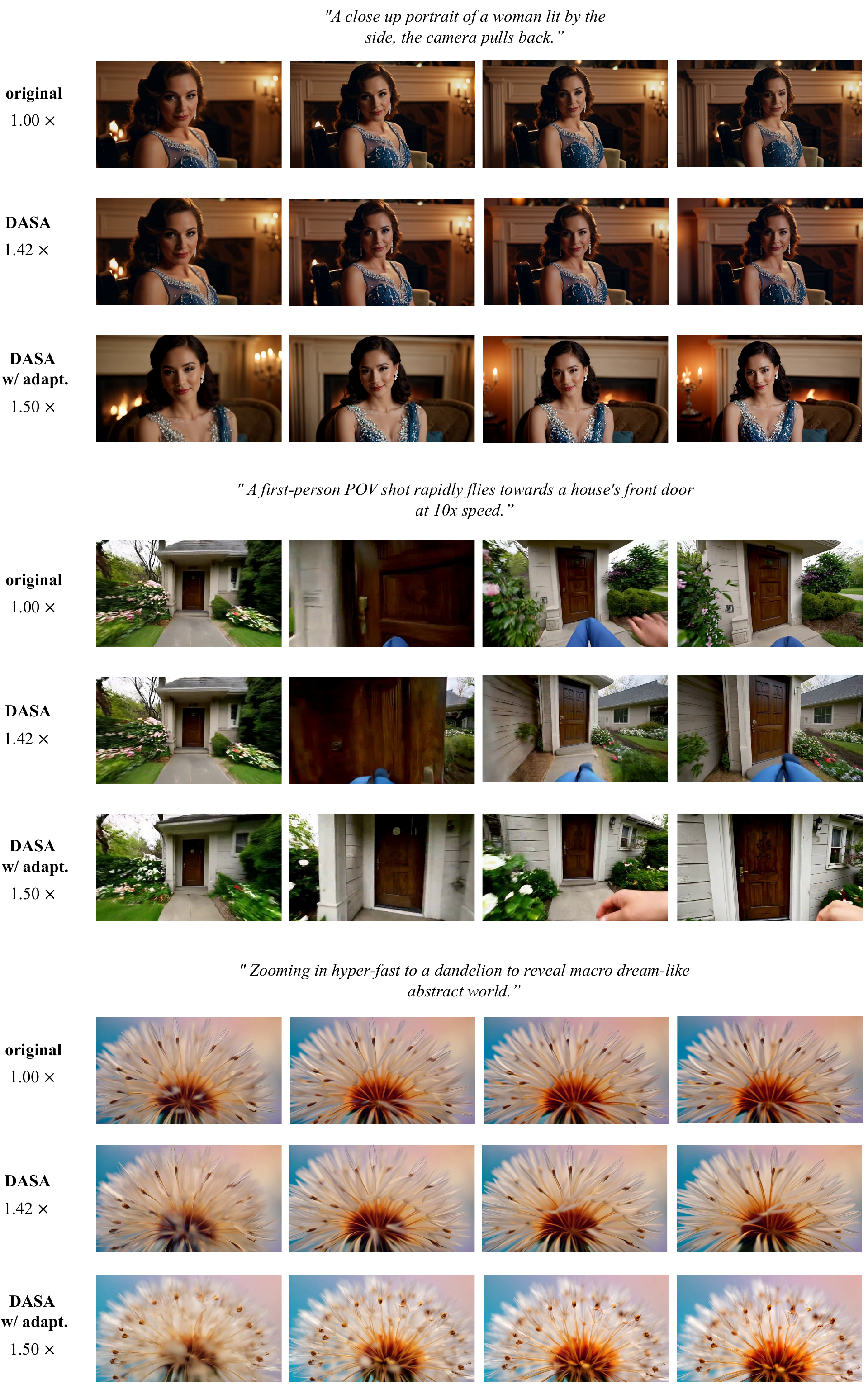}
  \caption{Visualization of 720P videos generated by Self Forcing.}
  \label{fig:fish}
\end{figure}

\begin{figure}[H]
  \centering
  \includegraphics[width=\textwidth]{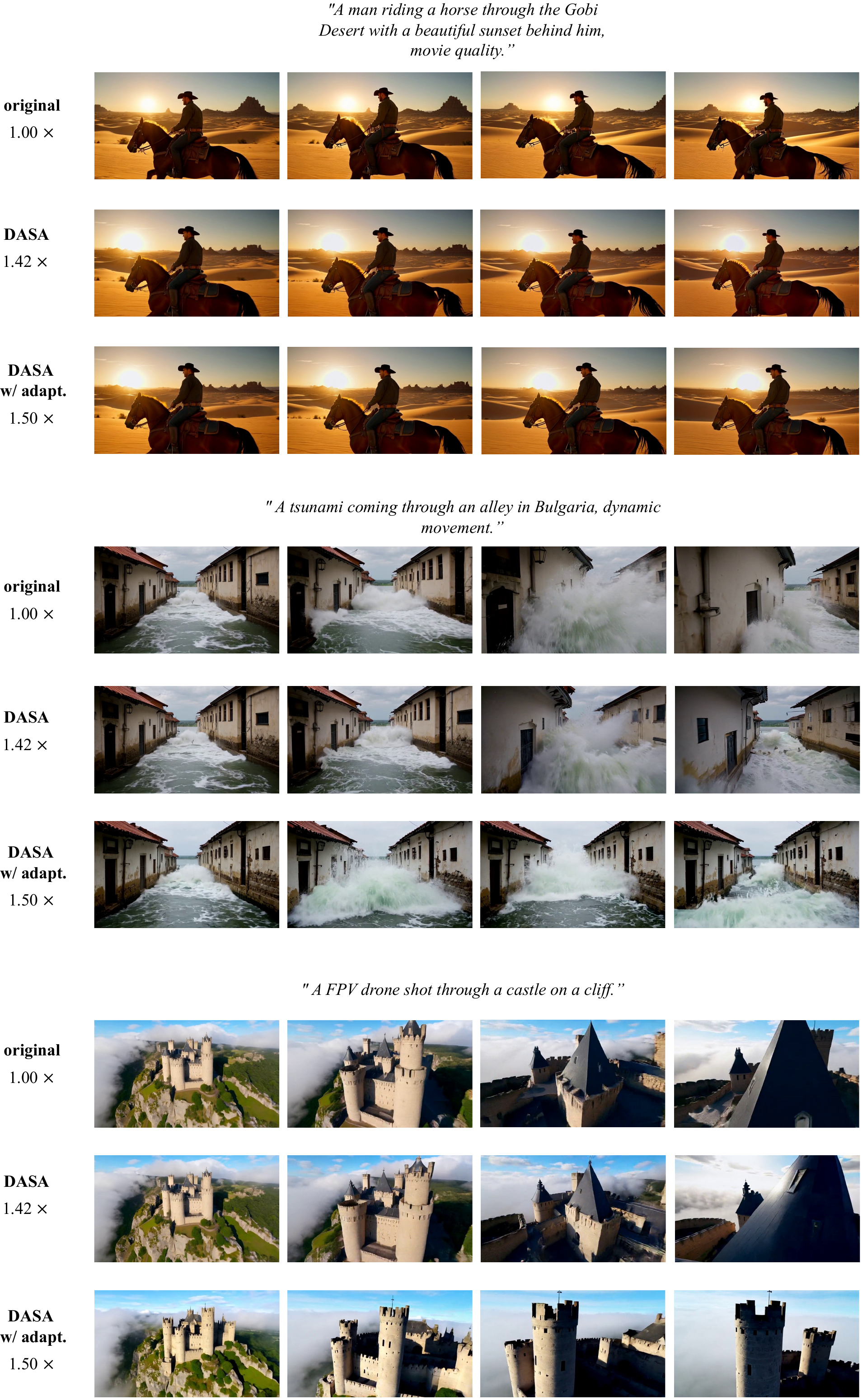}
  \caption{Visualization of 1080P videos generated by Self Forcing.}
  \label{fig:fish}
\end{figure}



\bibliographystyle{abbrv}
{
\small
\bibliography{reference}

@String(CVPR= {IEEE Conf. Comput. Vis. Pattern Recog.})

@String(ICCV= {Int. Conf. Comput. Vis.})

@String(NIPS= {Adv. Neural Inform. Process. Syst.})

@String(ICLR = {Int. Conf. Learn. Represent.})

@String(CVPR  = {CVPR})

@String(ICCV  = {ICCV})

@String(NIPS  = {NeurIPS})

@String{ICML  = {ICML}}

@String(ICLR  = {ICLR})

@article{wan2025wan,
  title={Wan: Open and advanced large-scale video generative models},
  author={Wan, Team and Wang, Ang and Ai, Baole and Wen, Bin and Mao, Chaojie and Xie, Chen-Wei and Chen, Di and Yu, Feiwu and Zhao, Haiming and Yang, Jianxiao and others},
  journal={arXiv preprint arXiv:2503.20314},
  year={2025}
}

@article{lin2024open,
  title={Open-sora plan: Open-source large video generation model},
  author={Lin, Bin and Ge, Yunyang and Cheng, Xinhua and Li, Zongjian and Zhu, Bin and Wang, Shaodong and He, Xianyi and Ye, Yang and Yuan, Shenghai and Chen, Liuhan and others},
  journal={arXiv preprint arXiv:2412.00131},
  year={2024}
}

@article{kong2024hunyuanvideo,
  title={Hunyuanvideo: A systematic framework for large video generative models},
  author={Kong, Weijie and Tian, Qi and Zhang, Zijian and Min, Rox and Dai, Zuozhuo and Zhou, Jin and Xiong, Jiangfeng and Li, Xin and Wu, Bo and Zhang, Jianwei and others},
  journal={arXiv preprint arXiv:2412.03603},
  year={2024}
}

@article{wu2025hunyuanvideo15,
  title={Hunyuanvideo 1.5 technical report},
  author={Wu, Bing and Zou, Chang and Li, Changlin and Huang, Duojun and Yang, Fang and Tan, Hao and Peng, Jack and Wu, Jianbing and Xiong, Jiangfeng and Jiang, Jie and others},
  journal={arXiv preprint arXiv:2511.18870},
  year={2025}
}

@article{seedance2026seedance2,
  title={Seedance 2.0: Advancing Video Generation for World Complexity},
  author={Seedance, Team and Chen, De and Chen, Liyang and Chen, Xin and Chen, Ying and Chen, Zhuo and Chen, Zhuowei and Cheng, Feng and Cheng, Tianheng and Cheng, Yufeng and others},
  journal={arXiv preprint arXiv:2604.14148},
  year={2026}
}

@article{seedance2025seedance15,
  title={Seedance 1.5 pro: A Native Audio-Visual Joint Generation Foundation Model},
  author={Seedance, Team and Chen, Heyi and Chen, Siyan and Chen, Xin and Chen, Yanfei and Chen, Ying and Chen, Zhuo and Cheng, Feng and Cheng, Tianheng and Cheng, Xinqi and others},
  journal={arXiv preprint arXiv:2512.13507},
  year={2025}
}

@online{kling25,
  author       = {{Kuaishou Technology}},
  title        = {Kling v2.5 Turbo: Video Generation Model},
  year         = 2025,
  url          = {https://klingai.com/global/},
  urldate      = {2026-01-13},
  note         = {Accessed: 2026-01-13.}
}

@misc{brooks2024sora,
  title={Video generation models as world simulators},
  author={Brooks, Tim and Peebles, Bill and Holmes, Connor and DePue, John and Guo, Yufei and Jing, Li and Schnurr, David and Taylor, Joe and Luhman, Troy and Luhman, Eric and others},
  year={2024},
  publisher={OpenAI},
  url={https://openai.com/research/video-generation-models-as-world-simulators},
  note={Accessed: 2026-05-02}
}

@article{lingbot-world,
      title={Advancing Open-source World Models}, 
      author={Robbyant Team and Zelin Gao and Qiuyu Wang and Yanhong Zeng and Jiapeng Zhu and Ka Leong Cheng and Yixuan Li and Hanlin Wang and Yinghao Xu and Shuailei Ma and Yihang Chen and Jie Liu and Yansong Cheng and Yao Yao and Jiayi Zhu and Yihao Meng and Kecheng Zheng and Qingyan Bai and Jingye Chen and Zehong Shen and Yue Yu and Xing Zhu and Yujun Shen and Hao Ouyang},
      journal={arXiv preprint arXiv:2601.20540},
      year={2026}
}

@article{hyworld2025,
  title={HY-World 1.5: A Systematic Framework for Interactive World Modeling with Real-Time Latency and Geometric Consistency},
  author={Team HunyuanWorld},
  journal={arXiv preprint},
  year={2025}
}

@article{wang2026matrix,
  title={Matrix-Game 3.0: Real-Time and Streaming Interactive World Model with Long-Horizon Memory},
  author={Wang, Zile and Liu, Zexiang and Li, Jaixing and Huang, Kaichen and Xu, Baixin and Kang, Fei and An, Mengyin and Wang, Peiyu and Jiang, Biao and Wei, Yichen and others},
  journal={arXiv preprint arXiv:2604.08995},
  year={2026}
}

@inproceedings{bruce2024genie,
  title={Genie: Generative interactive environments},
  author={Bruce, Jake and Dennis, Michael D and Edwards, Ashley and Parker-Holder, Jack and Shi, Yuge and Hughes, Edward and Lai, Matthew and Mavalankar, Aditi and Steigerwald, Richie and Apps, Chris and others},
  booktitle=ICML,
  year={2024}
}

@inproceedings{zhang2025framepack,
    title={Frame Context Packing and Drift Prevention in Next-Frame-Prediction Video Diffusion Models},
    author={Lvmin Zhang and Shengqu Cai and Muyang Li and Gordon Wetzstein and Maneesh Agrawala},
    booktitle=NIPS,
    year={2025},
}

@misc{ai2025magi1autoregressivevideogeneration,
      title={MAGI-1: Autoregressive Video Generation at Scale},
      author={Sand. ai and Hansi Teng and Hongyu Jia and Lei Sun and Lingzhi Li and Maolin Li and Mingqiu Tang and Shuai Han and Tianning Zhang and W. Q. Zhang and Weifeng Luo and Xiaoyang Kang and Yuchen Sun and Yue Cao and Yunpeng Huang and Yutong Lin and Yuxin Fang and Zewei Tao and Zheng Zhang and Zhongshu Wang and Zixun Liu and Dai Shi and Guoli Su and Hanwen Sun and Hong Pan and Jie Wang and Jiexin Sheng and Min Cui and Min Hu and Ming Yan and Shucheng Yin and Siran Zhang and Tingting Liu and Xianping Yin and Xiaoyu Yang and Xin Song and Xuan Hu and Yankai Zhang and Yuqiao Li},
      year={2025},
      eprint={2505.13211},
      archivePrefix={arXiv},
      primaryClass={cs.CV},
      url={https://arxiv.org/abs/2505.13211},
}

@inproceedings{yin2025causvid,
    title={From Slow Bidirectional to Fast Autoregressive Video Diffusion Models},
    author={Yin, Tianwei and Zhang, Qiang and Zhang, Richard and Freeman, William T and Durand, Fredo and Shechtman, Eli and Huang, Xun},
    booktitle=CVPR,
    year={2025}
}

@inproceedings{yin2024improved,
    title={Improved Distribution Matching Distillation for Fast Image Synthesis},
    author={Yin, Tianwei and Gharbi, Micha{\"e}l and Park, Taesung and Zhang, Richard and Shechtman, Eli and Durand, Fredo and Freeman, William T},
    booktitle=NIPS,
    year={2024}
}

@article{jiang2025lovic,
  title={Lovic: Efficient long video generation with context compression},
  author={Jiang, Jiaxiu and Li, Wenbo and Ren, Jingjing and Qiu, Yuping and Guo, Yong and Xu, Xiaogang and Wu, Han and Zuo, Wangmeng},
  journal={arXiv preprint arXiv:2507.12952},
  year={2025}
}

@article{huang2025self,
  title={Self forcing: Bridging the train-test gap in autoregressive video diffusion},
  author={Huang, Xun and Li, Zhengqi and He, Guande and Zhou, Mingyuan and Shechtman, Eli},
  journal={arXiv preprint arXiv:2506.08009},
  year={2025}
}

@article{zhu2026causal,
  title={Causal Forcing: Autoregressive Diffusion Distillation Done Right for High-Quality Real-Time Interactive Video Generation},
  author={Zhu, Hongzhou and Zhao, Min and He, Guande and Su, Hang and Li, Chongxuan and Zhu, Jun},
  journal={arXiv preprint arXiv:2602.02214},
  year={2026}
}

@article{liu2025rolling,
  title={Rolling forcing: Autoregressive long video diffusion in real time},
  author={Liu, Kunhao and Hu, Wenbo and Xu, Jiale and Shan, Ying and Lu, Shijian},
  journal={arXiv preprint arXiv:2509.25161},
  year={2025}
}

@article{zheng2024open,
  title={Open-sora: Democratizing efficient video production for all},
  author={Zheng, Zangwei and Peng, Xiangyu and Yang, Tianji and Shen, Chenhui and Li, Shenggui and Liu, Hongxin and Zhou, Yukun and Li, Tianyi and You, Yang},
  journal={arXiv preprint arXiv:2412.20404},
  year={2024}
}

@article{hong2022cogvideo,
  title={Cogvideo: Large-scale pretraining for text-to-video generation via transformers},
  author={Hong, Wenyi and Ding, Ming and Zheng, Wendi and Liu, Xinghan and Tang, Jie},
  journal={arXiv preprint arXiv:2205.15868},
  year={2022}
}

@inproceedings{gao2025ca2,
  title={Ca2-VDM: Efficient Autoregressive Video Diffusion Model with Causal Generation and Cache Sharing},
  author={Gao, Kaifeng and Shi, Jiaxin and Zhang, Hanwang and Wang, Chunping and Xiao, Jun and Chen, Long},
  booktitle=ICML,
  year={2025},
  organization={PMLR}
}

@article{gu2025long,
    title={Long-Context Autoregressive Video Modeling with Next-Frame Prediction},
    author={Gu, Yuchao and Mao, weijia and Shou, Mike Zheng},
    journal={arXiv preprint arXiv:2503.19325},
    year={2025}
}

@article{yang2025longlive,
      title={LongLive: Real-time Interactive Long Video Generation},
      author={Shuai Yang and Wei Huang and Ruihang Chu and Yicheng Xiao and Yuyang Zhao and Xianbang Wang and Muyang Li and Enze Xie and Yingcong Chen and Yao Lu and Song Hanand Yukang Chen},
      year={2025},
      eprint={2509.22622},
      archivePrefix={arXiv},
      primaryClass={cs.CV}
}

@article{guo2026efficient,
  title={Efficient Autoregressive Video Diffusion with Dummy Head},
  author={Guo, Hang and Jia, Zhaoyang and Li, Jiahao and Li, Bin and Cai, Yuanhao and Wang, Jiangshan and Li, Yawei and Lu, Yan},
  journal={arXiv preprint arXiv:2601.20499},
  year={2026}
}

@article{lv2026light,
  title={Light Forcing: Accelerating Autoregressive Video Diffusion via Sparse Attention},
  author={Lv, Chengtao and Shi, Yumeng and Huang, Yushi and Gong, Ruihao and Ren, Shen and Wang, Wenya},
  journal={arXiv preprint arXiv:2602.04789},
  year={2026}
}

@article{xu2026sparse,
  title={Sparse Forcing: Native Trainable Sparse Attention for Real-time Autoregressive Diffusion Video Generation},
  author={Xu, Boxun and Du, Yuming and Liu, Zichang and Yang, Siyu and Jiang, Ziyang and Yan, Siqi and Saha, Rajasi and Pumarola, Albert and Wang, Wenchen and Li, Peng},
  journal={arXiv preprint arXiv:2604.21221},
  year={2026}
}

@inproceedings{huang2024vbench,
  title={Vbench: Comprehensive benchmark suite for video generative models},
  author={Huang, Ziqi and He, Yinan and Yu, Jiashuo and Zhang, Fan and Si, Chenyang and Jiang, Yuming and Zhang, Yuanhan and Wu, Tianxing and Jin, Qingyang and Chanpaisit, Nattapol and others},
  booktitle=CVPR,
  pages={21807--21818},
  year={2024}
}

@article{HaCohen2024LTXVideo,
  title={LTX-Video: Realtime Video Latent Diffusion},
  author={HaCohen, Yoav and Chiprut, Nisan and Brazowski, Benny and Shalem, Daniel and Moshe, Dudu and Richardson, Eitan and Levin, Eran and Shiran, Guy and Zabari, Nir and Gordon, Ori and Panet, Poriya and Weissbuch, Sapir and Kulikov, Victor and Bitterman, Yaki and Melumian, Zeev and Bibi, Ofir},
  journal={arXiv preprint arXiv:2501.00103},
  year={2024}
}

@article{yang2024cogvideox,
  title={Cogvideox: Text-to-video diffusion models with an expert transformer},
  author={Yang, Zhuoyi and Teng, Jiayan and Zheng, Wendi and Ding, Ming and Huang, Shiyu and Xu, Jiazheng and Yang, Yuanming and Hong, Wenyi and Zhang, Xiaohan and Feng, Guanyu and others},
  journal={arXiv preprint arXiv:2408.06072},
  year={2024}
}

@article{wan2025,
      title={Wan: Open and Advanced Large-Scale Video Generative Models}, 
      author={Team Wan and Ang Wang and Baole Ai and Bin Wen and Chaojie Mao and Chen-Wei Xie and Di Chen and Feiwu Yu and Haiming Zhao and Jianxiao Yang and Jianyuan Zeng and Jiayu Wang and Jingfeng Zhang and Jingren Zhou and Jinkai Wang and Jixuan Chen and Kai Zhu and Kang Zhao and Keyu Yan and Lianghua Huang and Mengyang Feng and Ningyi Zhang and Pandeng Li and Pingyu Wu and Ruihang Chu and Ruili Feng and Shiwei Zhang and Siyang Sun and Tao Fang and Tianxing Wang and Tianyi Gui and Tingyu Weng and Tong Shen and Wei Lin and Wei Wang and Wei Wang and Wenmeng Zhou and Wente Wang and Wenting Shen and Wenyuan Yu and Xianzhong Shi and Xiaoming Huang and Xin Xu and Yan Kou and Yangyu Lv and Yifei Li and Yijing Liu and Yiming Wang and Yingya Zhang and Yitong Huang and Yong Li and You Wu and Yu Liu and Yulin Pan and Yun Zheng and Yuntao Hong and Yupeng Shi and Yutong Feng and Zeyinzi Jiang and Zhen Han and Zhi-Fan Wu and Ziyu Liu},
      journal = {arXiv preprint arXiv:2503.20314},
      year={2025}
}

@misc{chen2025skyreelsv2infinitelengthfilmgenerative,
      title={SkyReels-V2: Infinite-length Film Generative Model}, 
      author={Guibin Chen and Dixuan Lin and Jiangping Yang and Chunze Lin and Junchen Zhu and Mingyuan Fan and Hao Zhang and Sheng Chen and Zheng Chen and Chengcheng Ma and Weiming Xiong and Wei Wang and Nuo Pang and Kang Kang and Zhiheng Xu and Yuzhe Jin and Yupeng Liang and Yubing Song and Peng Zhao and Boyuan Xu and Di Qiu and Debang Li and Zhengcong Fei and Yang Li and Yahui Zhou},
      year={2025},
      eprint={2504.13074},
      archivePrefix={arXiv},
      primaryClass={cs.CV},
      url={https://arxiv.org/abs/2504.13074}, 
}

@article{deng2024nova,
  title={Autoregressive Video Generation without Vector Quantization},
  author={Deng, Haoge and Pan, Ting and Diao, Haiwen and Luo, Zhengxiong and Cui, Yufeng and Lu, Huchuan and Shan, Shiguang and Qi, Yonggang and Wang, Xinlong},
  journal={arXiv preprint arXiv:2412.14169},
  year={2024}
}

@article{jin2024pyramidal,
  title={Pyramidal Flow Matching for Efficient Video Generative Modeling},
  author={Jin, Yang and Sun, Zhicheng and Li, Ningyuan and Xu, Kun and Xu, Kun and Jiang, Hao and Zhuang, Nan and Huang, Quzhe and Song, Yang and Mu, Yadong and Lin, Zhouchen},
  jounal={arXiv preprint arXiv:2410.05954},
  year={2024}
}

@inproceedings{xie2025progressive,
  title={Progressive autoregressive video diffusion models},
  author={Xie, Desai and Xu, Zhan and Hong, Yicong and Tan, Hao and Liu, Difan and Liu, Feng and Kaufman, Arie and Zhou, Yang},
  booktitle=CVPR,
  pages={6322--6332},
  year={2025}
}

@inproceedings{zhai2025stargen,
  title={Stargen: A spatiotemporal autoregression framework with video diffusion model for scalable and controllable scene generation},
  author={Zhai, Shangjin and Ye, Zhichao and Liu, Jialin and Xie, Weijian and Hu, Jiaqi and Peng, Zhen and Xue, Hua and Chen, Danpeng and Wang, Xiaomeng and Yang, Lei and others},
  booktitle=CVPR,
  pages={26822--26833},
  year={2025}
}

@inproceedings{weng2024art,
  title={Art-v: Auto-regressive text-to-video generation with diffusion models},
  author={Weng, Wenming and Feng, Ruoyu and Wang, Yanhui and Dai, Qi and Wang, Chunyu and Yin, Dacheng and Zhao, Zhiyuan and Qiu, Kai and Bao, Jianmin and Yuan, Yuhui and others},
  booktitle=CVPR,
  pages={7395--7405},
  year={2024}
}

@inproceedings{sun2025ar,
  title={Ar-diffusion: Asynchronous video generation with auto-regressive diffusion},
  author={Sun, Mingzhen and Wang, Weining and Li, Gen and Liu, Jiawei and Sun, Jiahui and Feng, Wanquan and Lao, Shanshan and Zhou, SiYu and He, Qian and Liu, Jing},
  booktitle=CVPR,
  pages={7364--7373},
  year={2025}
}

@article{ji2026videoar,
  title={VideoAR: Autoregressive Video Generation via Next-Frame \& Scale Prediction},
  author={Ji, Longbin and Liu, Xiaoxiong and Shang, Junyuan and Wang, Shuohuan and Sun, Yu and Wu, Hua and Wang, Haifeng},
  journal={arXiv preprint arXiv:2601.05966},
  year={2026}
}

@article{huang2026live,
  title={LIVE: Long-horizon Interactive Video World Modeling},
  author={Huang, Junchao and Ye, Ziyang and Hu, Xinting and He, Tianyu and Zhang, Guiyu and Shi, Shaoshuai and Bian, Jiang and Jiang, Li},
  journal={arXiv preprint arXiv:2602.03747},
  year={2026}
}

@inproceedings{henschel2025streamingt2v,
  title={Streamingt2v: Consistent, dynamic, and extendable long video generation from text},
  author={Henschel, Roberto and Khachatryan, Levon and Poghosyan, Hayk and Hayrapetyan, Daniil and Tadevosyan, Vahram and Wang, Zhangyang and Navasardyan, Shant and Shi, Humphrey},
  booktitle=CVPR,
  pages={2568--2577},
  year={2025}
}

@inproceedings{chen2023seine,
  title={Seine: Short-to-long video diffusion model for generative transition and prediction},
  author={Chen, Xinyuan and Wang, Yaohui and Zhang, Lingjun and Zhuang, Shaobin and Ma, Xin and Yu, Jiashuo and Wang, Yali and Lin, Dahua and Qiao, Yu and Liu, Ziwei},
  booktitle=ICLR,
  year={2023}
}

@article{xi2025sparse,
  title={Sparse videogen: Accelerating video diffusion transformers with spatial-temporal sparsity},
  author={Xi, Haocheng and Yang, Shuo and Zhao, Yilong and Xu, Chenfeng and Li, Muyang and Li, Xiuyu and Lin, Yujun and Cai, Han and Zhang, Jintao and Li, Dacheng and others},
  journal={arXiv preprint arXiv:2502.01776},
  year={2025}
}

@article{zhang2025fast,
  title={Fast video generation with sliding tile attention},
  author={Zhang, Peiyuan and Chen, Yongqi and Su, Runlong and Ding, Hangliang and Stoica, Ion and Liu, Zhengzhong and Zhang, Hao},
  journal={arXiv preprint arXiv:2502.04507},
  year={2025}
}

@article{zhang2025spargeattention,
  title={Spargeattention: Accurate and training-free sparse attention accelerating any model inference},
  author={Zhang, Jintao and Xiang, Chendong and Huang, Haofeng and Wei, Jia and Xi, Haocheng and Zhu, Jun and Chen, Jianfei},
  journal={arXiv preprint arXiv:2502.18137},
  year={2025}
}

@article{yang2025sparse,
  title={Sparse videogen2: Accelerate video generation with sparse attention via semantic-aware permutation},
  author={Yang, Shuo and Xi, Haocheng and Zhao, Yilong and Li, Muyang and Zhang, Jintao and Cai, Han and Lin, Yujun and Li, Xiuyu and Xu, Chenfeng and Peng, Kelly and others},
  journal={arXiv preprint arXiv:2505.18875},
  year={2025}
}

@article{xu2025xattention,
  title={Xattention: Block sparse attention with antidiagonal scoring},
  author={Xu, Ruyi and Xiao, Guangxuan and Huang, Haofeng and Guo, Junxian and Han, Song},
  journal={arXiv preprint arXiv:2503.16428},
  year={2025}
}

@inproceedings{xia2025training,
  title={Training-free and adaptive sparse attention for efficient long video generation},
  author={Xia, Yifei and Ling, Suhan and Fu, Fangcheng and Wang, Yujie and Li, Huixia and Xiao, Xuefeng and Cui, Bin},
  booktitle=CVPR,
  pages={15982--15993},
  year={2025}
}

@article{zhang2025training,
  title={Training-free efficient video generation via dynamic token carving},
  author={Zhang, Yuechen and Xing, Jinbo and Xia, Bin and Liu, Shaoteng and Peng, Bohao and Tao, Xin and Wan, Pengfei and Lo, Eric and Jia, Jiaya},
  journal={arXiv preprint arXiv:2505.16864},
  year={2025}
}

@article{cui2026lol,
  title={LoL: Longer than Longer, Scaling Video Generation to Hour},
  author={Cui, Justin and Wu, Jie and Li, Ming and Yang, Tao and Li, Xiaojie and Wang, Rui and Bai, Andrew and Ban, Yuanhao and Hsieh, Cho-Jui},
  journal={arXiv preprint arXiv:2601.16914},
  year={2026}
}

@article{yuan2026helios,
  title={Helios: Real Real-Time Long Video Generation Model},
  author={Yuan, Shenghai and Yin, Yuanyang and Li, Zongjian and Huang, Xinwei and Yang, Xiao and Yuan, Li},
  journal={arXiv preprint arXiv:2603.04379},
  year={2026}
}

@misc{chen2026contextforcingconsistentautoregressive,
      title={Context Forcing: Consistent Autoregressive Video Generation with Long Context}, 
      author={Shuo Chen and Cong Wei and Sun Sun and Ping Nie and Kai Zhou and Ge Zhang and Ming-Hsuan Yang and Wenhu Chen},
      year={2026},
      eprint={2602.06028},
      archivePrefix={arXiv},
      primaryClass={cs.CV},
      url={https://arxiv.org/abs/2602.06028}, 
}

@inproceedings{zhang2025epona,
  title={Epona: Autoregressive diffusion world model for autonomous driving},
  author={Zhang, Kaiwen and Tang, Zhenyu and Hu, Xiaotao and Pan, Xingang and Guo, Xiaoyang and Liu, Yuan and Huang, Jingwei and Yuan, Li and Zhang, Qian and Long, Xiao-Xiao and others},
  booktitle=ICCV,
  pages={27220--27230},
  year={2025}
}

@article{yi2025deep,
  title={Deep forcing: Training-free long video generation with deep sink and participative compression},
  author={Yi, Jung and Jang, Wooseok and Cho, Paul Hyunbin and Nam, Jisu and Yoon, Heeji and Kim, Seungryong},
  journal={arXiv preprint arXiv:2512.05081},
  year={2025}
}

@article{chen2025sana,
  title={Sana-video: Efficient video generation with block linear diffusion transformer},
  author={Chen, Junsong and Zhao, Yuyang and Yu, Jincheng and Chu, Ruihang and Chen, Junyu and Yang, Shuai and Wang, Xianbang and Pan, Yicheng and Zhou, Daquan and Ling, Huan and others},
  journal={arXiv preprint arXiv:2509.24695},
  year={2025}
}

@inproceedings{liu2025timestep,
  title={Timestep Embedding Tells: It's Time to Cache for Video Diffusion Model},
  author={Liu, Feng and Zhang, Shiwei and Wang, Xiaofeng and Wei, Yujie and Qiu, Haonan and Zhao, Yuzhong and Zhang, Yingya and Ye, Qixiang and Wan, Fang},
  booktitle=CVPR,
  pages={7353--7363},
  year={2025}
}

@article{zhang2025dense,
  title={Dense video understanding with gated residual tokenization},
  author={Zhang, Haichao and Chai, Wenhao and He, Shwai and Li, Ang and Fu, Yun},
  journal={arXiv preprint arXiv:2509.14199},
  year={2025}
}

@article{sarkar2026cope,
  title={CoPE-VideoLM: Leveraging Codec Primitives For Efficient Video Language Modeling},
  author={Sarkar, Sayan Deb and Pautrat, R{\'e}mi and Miksik, Ondrej and Pollefeys, Marc and Armeni, Iro and Rad, Mahdi and Dusmanu, Mihai},
  journal={arXiv preprint arXiv:2602.13191},
  year={2026}
}

@inproceedings{yin2024one,
  title={One-step diffusion with distribution matching distillation},
  author={Yin, Tianwei and Gharbi, Micha{\"e}l and Zhang, Richard and Shechtman, Eli and Durand, Fredo and Freeman, William T and Park, Taesung},
  booktitle=CVPR,
  pages={6613--6623},
  year={2024}
}

@article{hong2025relic,
  title={Relic: Interactive video world model with long-horizon memory},
  author={Hong, Yicong and Mei, Yiqun and Ge, Chongjian and Xu, Yiran and Zhou, Yang and Bi, Sai and Hold-Geoffroy, Yannick and Roberts, Mike and Fisher, Matthew and Shechtman, Eli and others},
  journal={arXiv preprint arXiv:2512.04040},
  year={2025}
}

@article{li2025stable,
  title={Stable video infinity: Infinite-length video generation with error recycling},
  author={Li, Wuyang and Pan, Wentao and Luan, Po-Chien and Gao, Yang and Alahi, Alexandre},
  journal={arXiv preprint arXiv:2510.09212},
  year={2025}
}

@inproceedings{he2025neighboring,
  title={Neighboring autoregressive modeling for efficient visual generation},
  author={He, Yefei and He, Yuanyu and He, Shaoxuan and Chen, Feng and Zhou, Hong and Zhang, Kaipeng and Zhuang, Bohan},
  booktitle={2025 IEEE/CVF International Conference on Computer Vision (ICCV)},
  pages={19000--19010},
  year={2025},
  organization={IEEE}
}

@article{team2025inferix,
  title={Inferix: A Block-Diffusion based Next-Generation Inference Engine for World Simulation},
  author={Team, Inferix and Feng, Tianyu and Han, Yizeng and He, Jiahao and He, Yuanyu and Lin, Xi and Liu, Teng and Lu, Hanfeng and Tang, Jiasheng and Wang, Wei and others},
  journal={arXiv preprint arXiv:2511.20714},
  year={2025}
}

@inproceedings{he2026tc,
  title={TC-Pade: Trajectory-Consistent Pade Approximation for Diffusion Acceleration},
  author={He, Shaoxuan and Cui, Benlei and Huang, Bukun and Ye, Zhizeng and Sun, Yunyun and Huang, Longtao and Xue, Hui and Yang, Yang and Hong, Haiwen and Tang, Jingqun and others},
  booktitle={Proceedings of the IEEE/CVF Conference on Computer Vision and Pattern Recognition},
  pages={35768--35778},
  year={2026}
}

@misc{tang2026kforcingjointnextktokendecoding,
      title={K-Forcing: Joint Next-K-Token Decoding via Push-Forward Language Modeling}, 
      author={Zhiwei Tang and Yuanyu He and Yizheng Han and Wangbo Zhao and Jiasheng Tang and Fan Wang and Bohan Zhuang},
      year={2026},
      eprint={2606.10820},
      archivePrefix={arXiv},
      primaryClass={cs.LG},
      url={https://arxiv.org/abs/2606.10820}, 
}
}

\end{document}